\documentclass[journal]{IEEEtran}
\usepackage{verbatim}
\usepackage{amsfonts}

\usepackage{graphicx}
\usepackage{amsmath,amssymb} 
\usepackage{color}
\usepackage{epstopdf}

\usepackage[colorlinks,linkcolor=black]{hyperref}
\usepackage{multirow}
\usepackage{makecell}
\usepackage{color}
\usepackage{graphicx}

\usepackage{algorithm}
\usepackage{algorithmic}
\usepackage{ulem}

\ifCLASSINFOpdf
\else
\fi

\begin{document}
\title{SWIPENET: Object detection in noisy underwater images}
\author{
Long~Chen,~
Feixiang Zhou,
Shengke Wang,
Junyu Dong,~
Ning Li,~
Haiping Ma,~
Xin Wang~
and~Huiyu Zhou~
\thanks{L. Chen, F. Zhou and H. Zhou are with School of Informatics, University of Leicester, United Kingdom. H. Zhou is the corresponding author (Email: hz143@leicester.ac.uk). }
\thanks{N. Li is with College of Electronic and Information Engineering, Nanjing University of Aeronautics and Astronautics, China. }
\thanks{H. Ma is with Department of Electrical Engineering, Shaoxing University, China.
}
\thanks{S. Wang and J. Dong are with Department of information science and engineering, Ocean University of China, China. }
\thanks{X. Wang is with Colleage of Computer and Information, Hohai University, China. }
\thanks{Manuscript received on 1st Oct., 2020; revised xxxxx.}}


\maketitle
\begin{abstract}
In recent years, deep learning based object detection methods have achieved promising performance in controlled environments. However, these methods lack sufficient capabilities to handle underwater object detection due to these  challenges: (1) images in the underwater datasets and real applications are blurry whilst accompanying severe noise that confuses the detectors and (2) objects in real applications are usually small. In this paper, we propose a novel Sample-WeIghted hyPEr Network (SWIPENET), and a robust training paradigm named Curriculum Multi-Class Adaboost (CMA), to address these two problems at the same time. Firstly, the backbone of SWIPENET produces multiple high resolution and semantic-rich Hyper Feature Maps, which significantly improve small object detection. Secondly, a novel sample-weighted detection loss function is designed for SWIPENET, which focuses on learning high weight samples and ignore learning low weight samples. Moreover, inspired by the human education process that drives the learning from easy to hard concepts, we here propose the CMA training paradigm that first trains a 'clean' detector which is free from the influence of noisy data. Then, based on the 'clean' detector, multiple detectors focusing on learning diverse noisy data are trained and incorporated into a unified deep ensemble of strong noise immunity. \textcolor{black}{Experiments on four underwater object detection datasets show that the proposed SWIPENET+CMA framework achieves better or competitive accuracy in object detection against several state-of-the-art approaches.}
\end{abstract}
%
\IEEEpeerreviewmaketitle
\section{Introduction}
Autonomous underwater vehicles (AUVs) \cite{b1,b2} and remotely operated vehicles (ROVs) \cite{b3,b4} equipped with intelligent underwater object detection systems is of great significance for ocean resource exploitation and protection. Unfortunately, complicated underwater environments and lighting conditions introduce considerable noise into the captured images, which has posed massive challenges to intelligent vision-based object detection systems \cite{b5,b6,b7}. Therefore, it is crucial to develop novel underwater object detection techniques which effectively handle noise for the AUVs and ROVs applications.

\begin{figure}[tbp]
\centerline{\includegraphics[width=9cm, height=5cm]{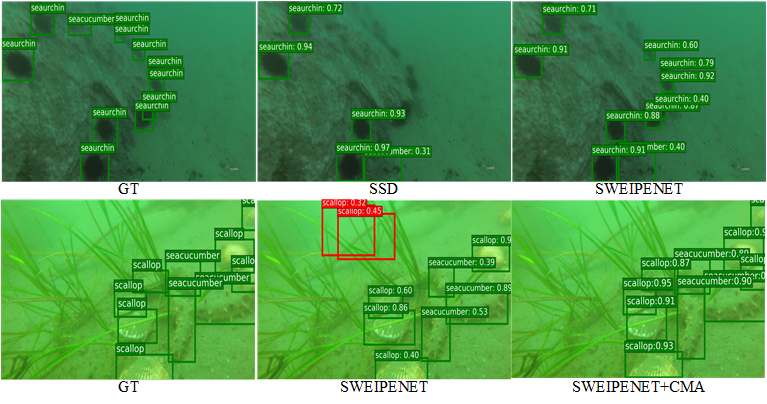}}
\caption{Exemplar images with ground truth (GT) annotations, results of Single Shot MultiBox Detector (SSD) \cite{b8}, our proposed SWIPENET and SWIPENET+CMA. The top row shows that SSD cannot detect all the small objects while our proposed SWIPENET outperforms SSD in this case. The bottom row shows our proposed SWIPENET treats the background as objects due to the existence of noisy data while our proposed SWIPENET+CMA performs better than the others.}
\label{fig:effectiveness}
\end{figure}
Deep learning based object detection systems have demonstrated promising performance in various applications but still felt short of dealing with underwater object detection. This is because, firstly, underwater detection datasets are scarce and the objects in the available underwater datasets and real applications are usually small. Current deep learning based detectors cannot effectively detect small objects (see an example shown on top row of Fig.~\ref{fig:effectiveness}). Secondly, the images in the existing underwater datasets and real applications accompany considerable noisy data. \textcolor{black}{In the underwater scenes, wavelength-dependent absorption and scattering \cite{b9} cause serious visibility loss, contrast decrease and color distortion, generating considerable noisy data. The noisy data exaggerates the challenge of inter-class similarity exist in object detection \cite{b01, b02}, resulting in the confusion between the object classes and the background class.} As shown on the bottom row of Fig.~\ref{fig:effectiveness}, the proposed SWIPENET trained on the noisy data cannot distinguish between the background and the objects.

In this paper, we propose a deep ensemble detector which is effective in dealing with small objects and noisy data in the underwater scenes. To achieve the objectives, we propose a deep backbone network named Sample-WeIghted hyPEr Network (SWIPENET), which fully takes advantage of multiple Hyper Feature Maps. To address the noisy data problem, we propose a novel sample-weighted detection loss function and a novel noise-robust training paradigm named Curriculum Multi-Class Adaboost (CMA), used to train the deep ensemble for underwater object detection. Indeed, the sample-weighted detection loss is used to control the influence of the training samples on SWIPENET. It works with the training paradigm CMA to train the proposed deep ensemble detector to reduce errors.

The proposed CMA training paradigm is inspired by the idea in the human education system that starts from learning easy tasks, and then gradually increase learning difficulty levels. This learning concept has been utilised to improve the generalisation ability and accelerate convergence in  machine learning algorithms \cite{b10, b11, b12}. For example, Derenyi et al. \cite{b12} reported theoretical analysis where easy examples should be learnt first due to less noise. They treat the samples misclassified by the Bayesian classifier as noisy data and learn the easier samples first, then improve convergence and the generalisation ability. Motivated by these works, our CMA training paradigm consists of two training stages: Noise-eliminating (NECMA) and noise-learning (NLCMA) stages. In the noise-eliminating stage, a 'clean' detector (SWIPENET) of being free from the influence of noisy data is formulated by focusing on learning easy samples whilst ignoring learning hard and noisy data. Then, the previously learnt knowledge by the 'clean' detector is again used to ease the training of the detectors in the noise-learning stage which focuses on learning diverse noisy data. The parameters of the detectors in the noise-learning stage are initialised by those of the 'clean' detector, which help the deep detectors avoiding poor local optimum during training and  improving the convergence speed and system generalisation. Finally, to achieve a balance between running time and detection accuracy, we present a selective ensemble algorithm to choose several detectors with a large diversity for the final ensemble. In summary, our contributions can be summarised as follows:
\begin{itemize}
\item We propose a novel noise-immune deep detection framework which consists of a backbone network SWIPENET and a powerful training paradigm CMA. The SWIPENET+CMA framework trains a robust deep ensemble detector for the object detection task in the underwater scenes with heterogeneous noisy data and small objects.
\item SWIPENET fully takes advantage of both high resolution and semantic-rich Hyper Feature Maps that significantly boost small object detection. Moreover, a novel sample-weighted detection loss is designed for the proposed SWIPENET, which controls the influence of the training samples on SWIPENET according to their weights. We also provide theoretical analysis on the ability of the sample-weighted detection loss in detail.
\item To achieve the balance between the detection accuracy and the computational cost, we propose a selective ensemble algorithm to choose the best detector trained with large data diversity.
\end{itemize}

The rest of the paper is organised as follows. Section \ref{sec:relatework} gives a brief introduction about the related work. Section \ref{sec:proposedmethod} describes our proposed SWIPENET backbone, CMA training paradigm and selective ensemble algorithm. Section \ref{sec:exmperiments} describes the experimental set-up and Section \ref{sec:results} reports the results of the proposed method on \textcolor{black}{four underwater object detection datasets.}

\section{Related Work}
\label{sec:relatework}
\subsection{Underwater object detection}
Underwater object detection techniques have been employed in marine ecology studies for many years. Strachan et al. \cite{b13} used color and shape descriptors to recognise fish transported on a conveyor belt, monitored by a digital camera. Spampinato et al. \cite{b14} presented a vision system for detecting, tracking and counting fish in real-time videos, which consist of video texture analysis, object detection and tracking processes. However, these established methods heavily rely on hand-crafted features, which have a limited representation ability. Villon et al. \cite{b16} compared a deep learning method against the Histogram of Oriented Gradients (HOG)+Support Vector Machine (SVM) method in detecting coral reef fish, and the experimental results show the superiority of the deep learning methods in underwater object detection. Li et al. \cite{b17, b19} exploited Fast RCNN \cite{b18} and Faster RCNN \cite{b20} to detect and recognise fish species. However, these methods use the features from the last convolution layer of the neural network, which is coarse and cannot effectively detect small objects. \textcolor{black}{To boost the feature representation capabilities, Fan et al. \cite{b54} proposed a stronger backbone named FERNet to exploit multi-scale contextual features, and a anchor refinement module is also introduced to solve the problem of sample imbalance. To achieve the high efficiency, Wang et al. \cite{b55} proposed a lightweight network named UnderwaterNet, which incorporates a Multi-scale Contextual Features Fusion (MFF) block and a Multi-scale Blursampling (MBP) module to reduce the network parameters. In addition, the underwater object detection datasets are extremely scarce that hinders the development of  underwater object detection techniques. Inspired by \cite{b241}, Chen et al. \cite{b42} proposed an Invert Multi-Class Adboost algorithm to down-weight the possible the noisy data and train the detection network, which and achieve state-of-the art performance on URPC2017 and URPC2018. Recently, Jian et al. \cite{b21, b22} proposed the OUC underwater dataset for underwater saliency detection with object-level annotations that can be used to evaluate the exiting systems. Lin et al. \cite{b56} proposed a data augmentation method RoIMix that focuses on interactions between images and mixes proposals among multiple images, this proposal-level data augmentation strategy greatly improve the performance of underwater object detectors.}

\subsection{Sample re-weighting}
Sample re-weighting is widely used to address noisy data problems \cite{b24} or hard sample mining \cite{b25}. It usually assigns a weight to each sample and then optimises the sample-weighted training loss. These can be divided into training loss based and results based methods. For the training loss based sample re-weighting approaches, we may have two research directions. For example, focal loss \cite{b26} and hard example mining \cite{b25} emphasise on hard samples with high training losses while self-placed learning \cite{b27, b28} encourages learning easy samples with low losses. These two possible solutions take different assumptions over the training data. The first solution assumes that hard samples are informative samples and should be learned more, whilst the second one assumes that hard samples are prone to be disturbance or noise. Here, in the underwater object detection tasks, hard samples are probably not useful because they confuse the detector rather than help it. Different from the training loss based sample-reweight methods, Multi-Class Adaboost \cite{b29} re-weights the samples according to the classification results. This method focuses on learning misclassified samples by increasing their weights during the iteration. Similarly, we re-weight the samples based on the detection results. This method seems more intuitive and effective than the training loss based methods.

\subsection{Curriculum leaning paradigm}
In the human education system, it may confuse the learner if s/he directly learns the hard knowledge in the beginning. Instead, the beginner starts from learning easy knowledge while skipping disturbing hard knowledge. In such way, the learning exercise is efficient and effective \cite{b30, b31}. This idea is also widely used in many machine leaning algorithms. For example, curriculum learning \cite{b32} and self-pace learning \cite{b27, b28} are two representatives inspired by the idea of learning easier aspects of the task before moving into a difficult level. Both approaches have been reported to provide better generalisation for the used model. However, Curriculum learning requires the samples in the datasets to be ranked in the order of incremental difficulty levels, but preparing such datasets is not trivial at all in practice. Self-pace learning addresses the sample order issue by training the used model and ranking the samples according to the samples' loss values using the learned model. It assumes the samples with low loss values are easy samples. One major drawback of self-pace learning is that it does not incorporate prior knowledge into the learning and hence loose the generalisation ability. In addition, both methods only train a single model without considering its capacity to learn diverse data. The developed models may be over-fit on some samples and under-fit on other samples. \textcolor{black}{In our work, we combine the learning tricks from Curriculum Learning and Multi-Class Adaboost into a novel noise-immune training paradigm CMA, which dynamically trains multiple detectors on the samples with a large diversity and combines them into a unified noise-immune deep ensemble detector.}

\section{Proposed method}
\label{sec:proposedmethod}
\begin{figure*}[tbp]
\centerline{\includegraphics[width=18cm, height=7.5cm]{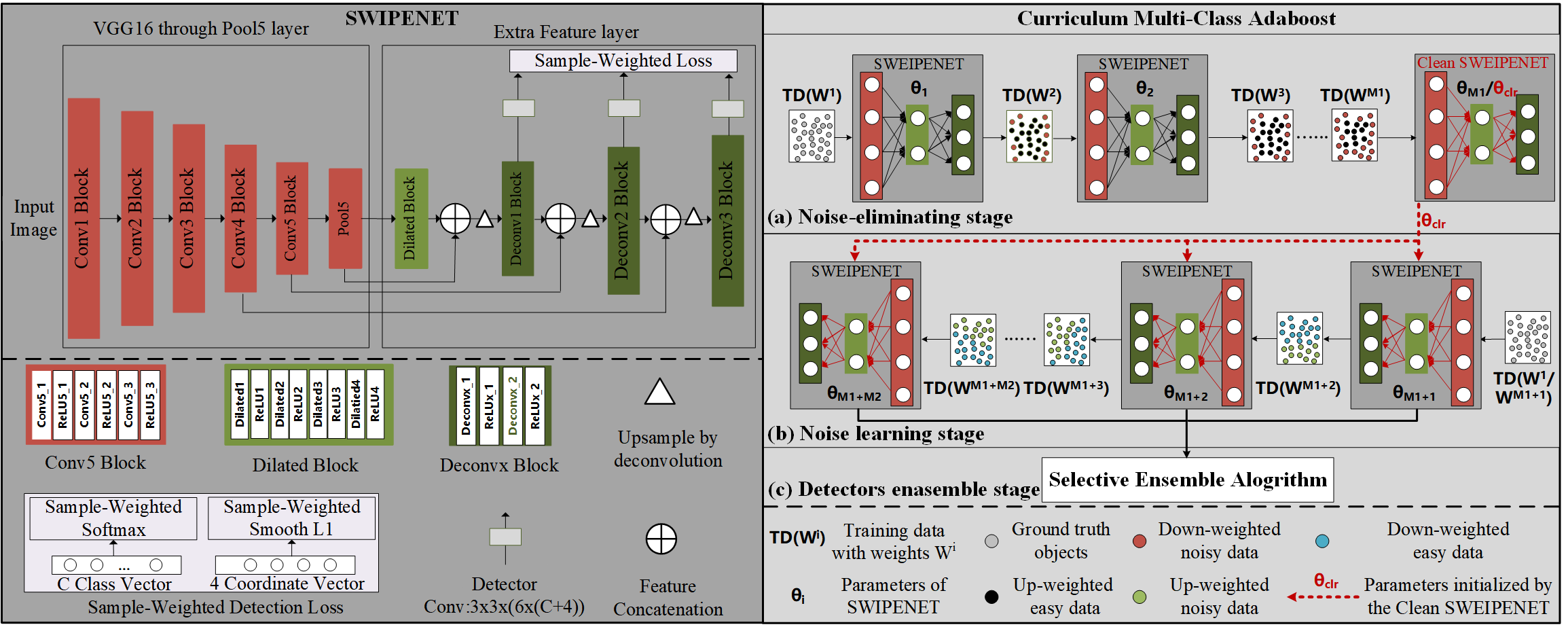}}
\caption{The overview of our proposed SWIPENET+CMA detection framework. The left shows the structure of our proposed SWIPENET backbone, the right shows the CMA training paradigm that consist of the \textcolor{black}{(a) Noise-eliminating stage (NECMA): gradually reduce the weights of the possible noisy data to obtain a ‘clean' detector which is free from the influence of the noisy data. (b) Noise learning stage (NLCMA): learn diverse noisy samples by increasing their weights to boost the generalisation ability. The parameters of each detector in NLCMA are initialised by those of the 'clean' detector, that alleviates the local optimum problem and accelerate the convergence, and (c) Detectors ensemble stage: ensemble multiple detectors to boost the generalisation ability.}}
\label{fig:SWIPENeT}
\end{figure*}
\subsection{Sample-WeIghted hyPEr Network (SWIPENET)}
Evidence shows that the down-sampling excises of Convolutional  Neural Network result in strong semantics that lead to the success of classification tasks. However, this is not enough for the object detection task which not only needs to recognise the objects but also spatially locates its position. After we have applied several down-sampling operations, the spatial resolutions of the deep layers are too coarse to handle small object detection.

\textcolor{black}{In this paper, we propose the SWIPENET architecture that includes several high resolution and semantic-rich Hyper Feature Maps inspired by Deconvolutional Single Shot Detector (DSSD) \cite{b33}, in which multiple down-sampling convolution layers are first constructed to extract high semantic feature maps. After several down-sampling operations, the feature maps are too coarse to provide sufficient information for accurate small object localization, therefore, multiple up-sampling deconvolution layers and skip connection are added to recover the high resolutions of the feature maps. However, the detailed information lost in the down-sampling operations cannot be fully recovered even though the resolutions have been recovered. Different from DSSD, we design a dilated convolution block in SWIPENET to obtain large receptive fields without sacrificing detailed information that support object localization (large receptive fields lead to strong semantics \cite{b34, b35}). The proposed dilated convolution block consists of 4 dilated convolution layers with ReLU activation and its detailed implementation can be found in Supplementary Section A.} Fig.~\ref{fig:SWIPENeT} illustrates the overview of our proposed SWIPENET, which consists of multiple convolution blocks, a novel dilated convolution block, multiple deconvolution blocks and a novel sample-weighted loss. The front layers of the SWIPENET are based on the architecture of the standard VGG16 model \cite{b36} (truncated at the Pool5 layer). \textcolor{black}{Then, we add the proposed dilated convolution block to extract high semantic while keep the resolutions of the feature maps. Finally, we up-sample the feature maps using deconvolution and add skip connection to construct multiple Hyper Feature Maps on the deconvolution layers.} The prediction of SWIPENET deploys three different deconvolution layers, i.e. Deconv1\_2, Deconv2\_2 and Deconv3\_2 (denoted as Deconvx\_2 in Fig.~\ref{fig:SWIPENeT}), which increase in size progressively and allow us to predict the objects of multiple scales. At each location of the three deconvolution layers, we define 6 default boxes and use a 3$\times$3 convolution kernel to produce $C+1$-D class prediction ($C$ indicates the number of the object classes and $1$ indicates the background class) and 4-D coordinate prediction. 

\subsection{Sample-Weighted detection loss}
We propose a novel sample-weighted detection loss function which can model sample weights in SWIPENET. The sample-weighted detection loss enables SWIPENET to focus on learning high weight samples whilst ignoring low weight samples. It cooperates with a novel sample re-weighting algorithm, namely Curriculum Multi-Class Adaboost, to reduce the influence of possible noise on SWIPENET by decreasing their weights.

Following the one-stage deep detector SSD \cite{b8}, SWIPENET trains an object detector using default boxes on several layers. \textcolor{black}{If the Intersection over Union (IoU) between the default box and its most overlapped object is larger than a pre-defined threshold, then the default box is a match to this object that works as its positive training sample. If a default box does not match any object, it will be regarded as a negative training sample.} Technically, our sample-weighted detection loss $L$ consists of a sample-weighted softmax loss $L_{cls}$ for the bounding box classification and a sample-weighted smooth L1 loss $L_{reg}$ for the bounding box regression:
\begin{equation} \small
L=\frac{\alpha_{1}}{\ddot{N}}L_{cls}(pre\_cls,gt\_cls,\bar{w})+\frac{\alpha_{2}}{\bar{N}} L_{reg}(pre\_loc,gt\_loc,\bar{w})
\label{formula:1}
\end{equation}
where $\ddot{N}$ and $\bar N$ are the numbers of all the training samples and positive training samples respectively, $\alpha_{1}$ and $\alpha_{2}$ denote the weight terms of classification and regression losses. The sample-weighted softmax loss $L_{cls}$ is formulated as
\begin{equation} \small
\begin{split}
L_{cls}=-\sum_{i=1}^{\ddot{N}} \sum_{c=1}^{C+1} \bar{w}_{i}^{m}gt\_cls_{i}^{c}log(pre\_cls_{i}^{c})
\label{formula:2}
\end{split}
\end{equation}

\begin{equation}\small
pre\_cls_{i}^{c}=\frac{e^{net_i^c}}{\sum_{\bar c=1}^{C+1}e^{net_i^{\bar c}}}
\label{formula:3}
\end{equation}
where $\bar{w}_i^m$ denotes the sample weight for the $i$-th sample computed in the $m$-th iteration of CMA in Subsection~
\ref{sec:cma}. \textcolor{black}{Denote $pre\_cls_i$ and $gt\_cls_i$ as the predicted and ground truth class vectors for the $i$-th sample, these two vectors are  $C+1$-D vectors ($C$ object classes plus one background class). $pre\_cls_{i}^{c}$ and $gt\_cls_{i}^{c}$ denote the $c$-th element of the predicted and ground truth class vectors for the $i$-th sample (referring to Supplementary Fig. 1 for better understanding).} $gt\_cls_i^c=1$ if the $i$-th sample belongs to the $c$-th class, $gt\_cls_i^c=0$ otherwise. $net_i^c$ is the classification prediction from the detection network.
$L_{reg}$ is the sample-weighted smooth L1 loss, formulated as follows:
\begin{equation}\small
L_{reg}=\sum_{i=1}^{\bar N}\sum_{l\in Loc}\bar{w}_{i}^{m}Smooth_{L_1}(pre\_loc_{i}^{l}-gt\_loc_{j}^{l})
\label{formula:reg}
\end{equation}
\begin{equation}\small
Smooth_{L_1}(x)=\left\{\begin{matrix} 0.5x^2 \qquad\ \textbf{if } |x|<1
\\
|x|-0.5 \quad \textbf{otherwise}
\end{matrix}\right.
\label{formula:smoothl1}
\end{equation}
\begin{equation}\small
pre\_loc_i^l=net_i^l, l \in Loc
\label{formula:smoothl1}
\end{equation}

\textcolor{black}{$pre\_loc_i$ and $gt\_loc_i$ denote the predicted and ground truth coordinate vectors for the $i$-th sample, these two vectors are  4-D vectors (the coordinate information $Loc=(cx,cy,w,h)$ includes the coordinates of center $(cx,cy)$ with width $w$ and height $h$.} $pre\_loc_{i}^{l}$ and $gt\_loc_{i}^{l}$ denote the $l$-th element of the predicted and the ground truth coordinate vectors for the $i$-th positive training sample respectively. $net_i^l$ is the coordinate prediction from the detection network.

\textcolor{black}{In the gradient based optimisation algorithm, the loss function plays a key role in providing the gradients for updating the model parameters in the back-propagation process. The sample's gradient magnitude in the derivative of the loss function determines its impact on the updating of the DNNs. In our proposed sample-weighted detection loss, the sample weight $\bar{w}$ is able to adjust the sample's gradient magnitude. Hence, we are able to investigate how the sample weight influences the sample's impact on the feature learning of DNNs. Denote the parameter of the detector as $\theta$, the derivative of the sample-weighted detection loss $\frac{\partial L}{\partial \theta}$ is derived as (the detailed derivation process can be found in the Supplementary Section. B):}
\begin{equation}\small
\begin{aligned}
\frac{\partial L}{\partial \theta}=&\left\{\begin{matrix} \frac{\alpha_1}{\ddot{N}}\sum_{i=1}^{\ddot{N}} \sum_{c=1}^{C+1}\bar{w}_{i}^{m}gt\_cls_{i}^{c} (pre\_cls_i^c-1) \frac{\partial net_i^c}{\partial \theta}\\
+ \frac{\alpha_2}{\bar N} \sum_{i=1}^{\bar N} \sum_{l \in Loc} \bar{w}_{i}^{m}(pre\_loc_{i}^{l}-gt\_loc_{j}^{l}) \frac{\partial net_i^l}{\partial \theta}\\
\quad\quad\quad\quad\quad\quad\quad\quad\quad\ \textbf{if } |pre\_loc_{i}^{l}-gt\_loc_{j}^{l}|<1
\\ \frac{\alpha_1}{\ddot{N}}\sum_{i=1}^{\ddot{N}} \sum_{c=1}^{C+1}\bar{w}_{i}^{m}gt\_cls_{i}^{c} (pre\_cls_i^c-1) \frac{\partial net_i^c}{\partial \theta}\\
\pm \frac{\alpha_2}{\bar N} \sum_{i=1}^{\bar N} \sum_{l \in Loc} \bar{w}_{i}^{m} \frac{\partial net_i^l}{\partial \theta} \qquad\qquad \textbf{otherwise}
\end{matrix}\right.\\
\end{aligned}
\label{eq:finalderive}
\end{equation}

\textcolor{black}{From Eq.~(\ref{eq:finalderive}), we witness that the sample's gradient magnitude in the derivative is influenced by two factors. The first one is the accuracy of the predicted class and coordinates. For the $i$-th training sample with ground truth class $c$ (i.e., $gt\_cls_i^c=1$), the closer $pre\_cls_i^c$ and $pre\_loc_i^c$ to the ground truth, the smaller the gradient magnitude for the $i$-th sample. Second, the sample's weight $\bar{w}_i^m$. Suppose all of the training samples have the same prediction accuracy. The smaller the weight is, the smaller gradient magnitude is attached to the $i$-th sample. For example, if we assign a weight of 100 and 1 to the same positive sample respectively, then the gradient magnitude of the former one may be around 100 times that of the later one. The feature learning of DNN is dominated by high-weight samples while the low-weight samples contribute far less to the update of the DNN features. Hence, the sample-weighted detection loss less counts on the low-weight samples.}

\subsection{Curriculum Multi-class Adaboost (CMA)}
\label{sec:cma}
Underwater images suffer from the degradation of different levels, e.g. poor lighting, noise and blurs. If we train a detector on the dataset containing severely deteriorated images, the 'noisy data' are easy to confuse the object detector. Such example is illustrated in the bottom of Fig.~\ref{fig:effectiveness}.

\textcolor{black}{Inspired by the human education system that learns from easy to hard samples, we here propose a noise-immune training paradigm, namely Curriculum Multi-class Adaboost (CMA), to train multiple deep detectors and then ensemble them into a unified model for underwater object detection in the underwater scenes with the data of considerable noise and large diversity.} This strategy helps accelerate the convergence and improve the generalisation of the proposd architecture because the detector trained on easy data provides optimum initialisation for the following deep detectors. Good initialisation helps the proposed model to avoid the local optimum problem in training and to improve generalisation, which has been demonstrated in previous work \cite{b10, b11, b12}.

\subsubsection{The overview of the CMA}
\textcolor{black}{CMA is developed based on Multi-Class Adaboost (MA) \cite{b29}, which trains multiple base classifiers sequentially and assign a weight value $\alpha_{m}$ to each classifier according to its error rate $E_{m}$. When training each classifier, the samples misclassified by the preceding classifier are assigned a higher weight, allowing the following classifier to focus on learning these samples. Finally, all the weak base classifiers are combined to form an ensemble classifier with corresponding weight values}.

Different form MA, our proposed CMA algorithm consists of two stages: noise-eliminating (denotes as NECMA) and noise-learning stages (denotes as NLCMA). In each training iteration of NECMA, we reduce the weights of the undetected objects as they are likely to be noisy data \cite{b12}. \textcolor{black}{The sample-weighted detection loss enables the next SWIPENET to only focus on learning the high-weight clean data. By gradually reducing the influence of the noisy data, the generalisation capability of the system is improved and a detector free from the influence of the noisy data is sought. However, after several iterations, the deep detector may over-fit over the clean, easy samples as their weights are too high after several rounds of re-weighting exercises, and the generalisation ability becomes deteriorated.} Therefore, we terminate the noise-eliminating stage when the performance does not improve anymore, and the detector achieving the best detection accuracy is selected as the 'clean' detector.

\textcolor{black}{The 'clean' detector can detect the clean, easy objects well but always fails to detect many hard objects. This is because although the undetected objects tend to be noisy data or outliers, but they also contain many hard object instances which haven't been detected because they are very similar with the backgrounds. Ignoring these hard object instances will limit the network's generalization ability on the hard object instance. Hence, we propose the NLCMA training stage, which focuses on learning diverse hard samples by increasing their weights. In practice, the parameters of each detector in NLCMA are initialised by those of the 'clean' detector. This strategy effectively alleviates the local optimum problem and significantly accelerate the convergence whilst boosting the generalisation ability.}

\begin{algorithm}
\caption{Noise-immune CMA training paradigm.}
\label{alg:1}
\textbf{Input}: Training images $I_{train}$ with ground truth objects $B=\{b_{1},...,b_{N}\}$, testing images $I_{test}$.\\
\textbf{Output}: $M$ SWIPENETs.
\begin{algorithmic}[1] 
\STATE Initialise the object weights $w_{j}^{1}=\frac{1}{N}, j=1,...,N$.
\FOR{$m=1$ to $M_1$}
\STATE $\bullet$  Compute the weights of positive samples using \textcolor{black}{Eq. (8)}.\\
\STATE $\bullet$  Train the $m$-th SWIPENET $G_{m}$ using Eq. (1).\\
\STATE $\bullet$ Compute the $m$-th SWIPENET's error rate $E_{m}$ using \textcolor{black}{Eqs. (9)-(10)}.
\STATE $\bullet$ Compute the $m$-th SWIPENET's weight $\alpha_{m}$ in the ensemble model using \textcolor{black}{Eq. (11)}.
\STATE $\bullet$ Reduce the weights of the undetected objects and increase the weights of the detected objects using \textcolor{black}{Eq. (12)}).
\ENDFOR
\STATE Obtain the parameter $\theta_{clr}$ of the $M_1$-th SWIPENET.
\STATE Initialize the object weights $w_{j}^{M_1+1}=\frac{1}{N}, j=1,...,N$.
\FOR{$m=M_1+1$ to $M_2$}
\STATE $\bullet$  Compute the weights of positive samples using \textcolor{black}{Eq. (8)}.\\
\STATE $\bullet$  Initialize the parameter of the $m$-th SWIPENET $G_{m}$ using $\theta_{base}$.\\
\STATE $\bullet$  Train the $m$-th SWIPENET $G_{m}$ using Eq. (1).\\
\STATE $\bullet$ Compute the $m$-th SWIPENET's error rate $E_{m}$ using \textcolor{black}{Eqs. (9)-(10)}.
\STATE $\bullet$ Compute the $m$-th SWIPENET's weight $\alpha_{m}$ in the ensemble model using \textcolor{black}{Eq. (11)}.
\STATE $\bullet$ Increase the weights of the undetected objects and decrease the weights of the detected objects using \textcolor{black}{Eq. (13)}.
\ENDFOR
\STATE \textbf{return} $M$ SWIPENETs.
\end{algorithmic}
\end{algorithm}
The proposed CMA training paradigm can be found in Algorithm 1. \textcolor{black}{It iteratively trains $M$ detectors, including $M_1$ iterations for NECMA and $M_2$ iterations for NLCMA. We assume the best performing detector (i.e, the 'clean' detector $S_{clr}$ parameterised by $\theta_{clr}$) in NECMA is achieved in the $M_1$-th iteration, $M_1$ is experimentally obtained.} Denote $I_{train}$ as the training images with the ground truth objects $B=\{b_{1}, b_{2},..., b_{N}\}$, $N$ is the number of the objects in the training set, $b_{j}=(cls,cx,cy,w,h)$ is the annotation of the $j$-th object. We denote $w_{j}^{m}$ as the weight of the $j$-th object in the $m$-th iteration. Each object's weight is initialised to $\frac{1}{N}$ in the first iteration, i.e. $w_{j}^{1}=\frac{1}{N}, j=1,...,N$.

In the $m$-th iteration of CMA, we firstly compute the weights of the positive training samples. If the $i$-th positive sample matches the $j$-th object during the training, we compute the $i$-th positive sample's weight $\bar{w}_{i}^{m}$ using Eq.~(\ref{eq:poswei}).
\begin{equation}
\bar{w}_{i}^{m} = N*w_{j}^{m}, 0<w_{j}^{m}<1
\label{eq:poswei}
\end{equation}
where $w_{j}^{m}$ denotes the weight of the $j$-th object in the $m$-th iteration. The weight of the positive sample is $N$ times that of its matched object. This is because the initial weight of each object in CMA is $\frac{1}{N}$, and the initial weight of each positive training sample in the sample-weighted detection loss is 1.  Secondly, we use the re-weighted samples to train the $m$-th detector $S_{m}$. Thirdly, we run the $m$-th detector on the training set and receive the detection results $D_{m}=\{d_{1}, d_{2},..., d_{i}\}$ while $d_{i}=(cls,score,cx,xy,w,h)$ is the $i$-th predicted outcome, including the predicted class ($cls$), score ($score$) and coordinates ($cx,cy,w,h$). \textcolor{black}{The error rate $E_{m}$ of the $m$-th detector is computed based on the percentage of the undetected objects.}
\begin{equation}\small
E_{m}=\sum_{j=1}^{N}w_{j}^{m}I(b_{j})/\sum_{j=1}^{N}w_{j}^{m}
\label{formula:5}
\end{equation}
where
\begin{equation}\small
    \displaystyle
    I(b_{j})=\left\{
             \begin{array}{lr}
             0\,\textbf{   if }\exists\, d\in D_{m},\, s.t. b_{j}.cls==d.cls \wedge IoU(b_{j},d)\geq \theta\, \\
             1\,\textbf{   otherwise} \\
             \end{array}
\right.
\label{formula:invertindicator}
\end{equation}
In Eq.~(\ref{formula:invertindicator}), if there exists a detection $d$ which belongs to the same class as the $j$-th ground truth object $b_{j}$ (i.e. $b_{j}.cls==d.cls$) and the Intersection over Union (IoU) between the detection and the $j$-th object is larger than the threshold $\theta$ (0.5 here), we set $I(b_{j})=0$, indicating the $j$-th object has been detected and $I(b_{j})=1$ is the undetected. Fourthly, we compute the $m$-th detector's weight $\alpha_{m}$ using Eq.~(\ref{eq:modelweight}), which is used when we ensemble different detectors.
\begin{equation}\small
\alpha_{m}=log\frac{1-E_{m}}{E_{m}}+log(C-1)
\label{eq:modelweight}
\end{equation}

\begin{equation}\small
w_{j}^{m} \leftarrow \frac{w_{j}^{m}}{z_{m}}exp(\alpha _{m}(1-I(b_{j})))
\label{formula:invertupdate}
\end{equation}
\textcolor{black}{where $C$ is the number of the object classes. Finally, we update each object's weight $w_{j}^{m}$ and train the following detector. In the first $M_1$ iterations of NECMA stage, we reduce the weights of the undetected objects by Eq.~(\ref{formula:invertupdate}) that enables the next detector to pay less attention to possible noisy data. In the last $M_2$ iterations of the NLCMA stage, we increase the weights of the undetected objects by Eq.~(\ref{formula:update}), whereas the detector turns to learning the diverse hard data.} $z_{m}$ is a normalisation constant. The same iteration repeats again till all $M$ detectors have been trained.

\begin{equation}\small
w_{j}^{m} \leftarrow \frac{w_{j}^{m}}{z_{m}}exp(\alpha _{m}I(b_{j}))
\label{formula:update}
\end{equation}
\textcolor{black}{It is noticed that when CMA changes from NECMA to NLCMA, i.e., in the $M_1+1$-th iteration, we must re-initialise the weight of each object as $\frac{1}{N}$.} In each iteration of NLCMA, we initialise the parameter of each detector with the parameter $\theta_{clr}$ of the 'clean' SWIPENET. These initialisations help the system avoid local maximum (or minimum) problem, whilst efficiently converging to stationary points.

\subsubsection{Selective ensemble algorithm}
An ensemble model may be more accurate than a single model, but brings in additional computational overhead. Recent references \cite{b37,b38,b39} have pointed out that the ensemble of selective deep models may not only be more compact but also stronger in the generalization ability than that of the overall deep models. \textcolor{black}{To reduce the computational costs, we first select a few detectors with large diversity. If the results of two different detectors look similar, the ensemble model based on the two detectors does not have the complementary ability. We need to determine which detector is used and how many detectors are incorporated in the final ensemble model.}

We here propose a greedy selection algorithm to select candidate detectors for the final ensemble. \textcolor{black}{Firstly, we construct a candidate ensemble set $E$ to add up the selected detectors, and initialise it with the detector achieving the highest detection accuracy among all the $M_2$ detectors in NLCMA as these detectors have not been confused by noisy data. Then, we gradually select a single detector $D_{m^{*}}$ having the largest diversity with all the detectors in the candidate ensemble set and add it to the ensemble set}, as formulated in Eq. (\ref{eq:Q}).

\begin{equation}
\begin{aligned}
D_{m^{*}} = \mathop{\arg\max}_{m,D_m \notin E} \sum_{D_n\in E}Q_{mn}
\end{aligned}
\label{eq:Q}
\end{equation}
Here, we apply the commonly used Q statistic \cite{b40} to measuring the diversity of two detectors' performance.
\begin{equation}
\begin{aligned}
Q_{mn}=\frac{N^{11}N^{00}-N^{01}N^{10}}{N^{11}N^{00}+N^{01}N^{10}}
\end{aligned}
\label{eq:22}
\end{equation}
$Q_{mn}$ denotes the diversity between the performance of detectors $D_m$ and $D_n$. $N^{11}$ and $N^{00}$ are the numbers of the objects detected and missed by the two detectors respectively. $N^{01}$ is the total number of the objects missed by $D_m$ and detected by $D_n$, $N^{10}$ is the total number of the objects detected by $D_m$ and missed by $D_n$. Maximum diversity is achieved at $Q_{mn}=-1$ when the two detectors make different predictions (i.e., $N^{11}=N^{00}=0$), and the minimum diversity is achieved at $Q_{mn}=1$ when the two detectors generate identical predictions (i.e., $N^{01}=N^{10}=0$).

\textcolor{black}{After all the candidate detectors have been selected, we ensemble them into a unified ensemble detector according to their weights computed by Eq. (\ref{eq:modelweight}) in CMA and their diversity weight in the ensemble set.} We assign a higher weight to the detector with a larger diversity. This enables the ensemble detector to detect diverse objects in the underwater scenes, where a large diversity exists due to the changed illuminations, water depth and object-camera distance. We compute the diversity weight $div_m$ of detector $D_m$ as its average diversity with all the detectors in the ensemble set (by Eq. (\ref{eq:div})).

\begin{equation}
\begin{aligned}
div_{m} = \sum_{D_n\in E, n \neq m}Q_{mn}^*/(|E|-1)
\end{aligned}
\label{eq:div}
\end{equation}
The value of $Q_{mn}$ lies in [-1,1]. For better representing the weights of the detection model, we normalise $Q_{mn}$ as $Q_{mn}^*$ using Eq.~(\ref{eq:Q*}). The value of $Q_{mn}^*$ lies in [0,1], and the larger diversity the large value of the diversity weight.

\begin{figure*}[h]
\centerline{\includegraphics[width=18cm, height=5.8cm]{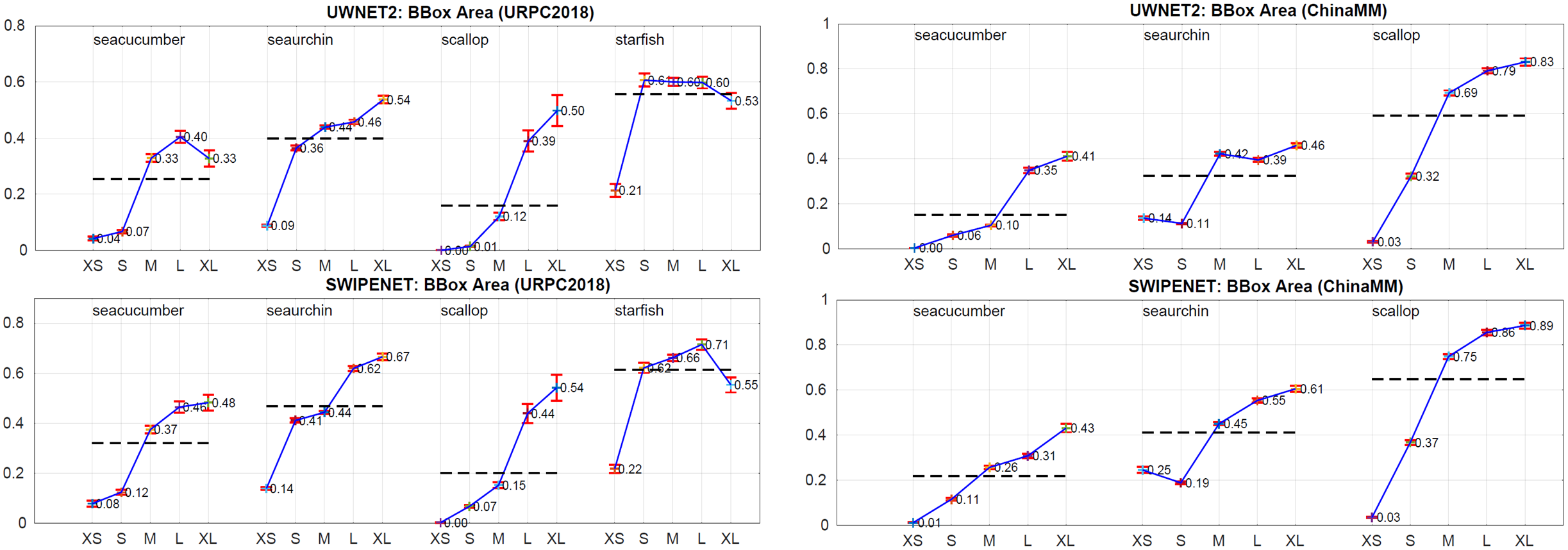}}
\caption{\textcolor{black}{The mean Average Precision of UWNET2 and SWIPENET for objects with different object sizes on URPC2018 and ChinaMM. The object size is measured as the pixel area of the bounding box. XS (bottom 10\%)=extra-small; S (next 20\%)=small; M (next 40\%)=medium; L (next 20\%)=large; XL (next 10\%)=extra-large.}}
\label{fig:ObjectSize}
\end{figure*}
\begin{equation}
\begin{aligned}
Q_{mn}^*=0.5(1-Q_{mn})
\end{aligned}
\label{eq:Q*}
\end{equation}
The final weight $\lambda_{i}$ of detector $D_i$ is formulated as
\begin{equation}
\begin{aligned}
\lambda_{i} = \frac{div_i*\alpha_i}{\sum_{m=1}^{M^*}div_m*\alpha_m}M^*, i=1,...,M^*
\end{aligned}
\label{eq:finalwei}
\end{equation}
In the testing stage, we use the weights to re-score the detection boxes. $M^*$ denotes the number of the selected detectors, and $M^*/\sum_{m=1}^{M^*}div_m*\alpha_m$ in Eq. (\ref{eq:finalwei}) is a normalisation term, scaling  the score of the box to fall in [0,1] after re-scoring. In particular, we first run all $M^*$ selected SWIPENETs on the testing set $I_{test}$ and produce a $M^*$ detection set $Det_m$.
\begin{equation}
\begin{aligned}
Det_m=D_{m}(I_{test}), m=1,2,...,M^*
\end{aligned}
\label{eq:Det_m}
\end{equation}
Afterwards, we re-score each detection $d$ in $Det_{m}$ using $\lambda_m$.
\begin{equation}
\begin{aligned}
d.score=\lambda_{m} d.score, d\in Det_m
\end{aligned}
\label{eq:rescore}
\end{equation}
\textcolor{black}{Finally, we combine all the detections and utilise Non-Maximum Suppression \cite{b41} to remove the overlapped detections by Eq.~(\ref{formula:NMS}), achieving the final detection results $Det$.}
\begin{equation}\small
Det = NonMaximumSuppression(\bigcup_{m=1}^{M^*} Det_{m})
\label{formula:NMS}
\end{equation}

\section{Experiments Setup}
\label{sec:exmperiments}
To demonstrate the effectiveness of the proposed method, \textcolor{black}{we conduct comprehensive evaluations on four underwater datasets URPC2017, URPC2018, URPC2019 and ChinaMM~\cite{b46}. The former three datasets come from the Underwater Robot Picking Contest (detailed descriptions of the contest are provided in the Supplementary Section. C).} In this section, we first introduce the experimental datasets. Then, we describe the implementation details.

\subsection{Datasets}
\textcolor{black}{The URPC2017 and ChinaMM datasets have 3 object categories, including seacucumber, seaurchin and scallop. URPC2017 contains 18,982 training images and 983 testing images. ChinaMM contains 2,071 training images and 676 validation images. The URPC2018 and URPC2019 datasets have 4 object categories, including seacucumber, seaurchin, scallop and starfish. URPC2018 and URPC2019 have published the training set, but the testing set is not publicly available. Hence, we randomly split the training set of URPC2018 into 1,999 training images and 898 testing images, and split the training set of URPC2019 into 3,409 training images and 1,000 testing images. All four datasets provide underwater images and box level annotations.}

\subsection{Implementation details}
All the experiments are conducted on a server with an Intel Xeon CPU @ 2.40GHz and a single Nvidia Tesla P100 GPUs with a 16 GB memory. For our proposed detection framework, we implement it using the Keras framework, and train it with the Adam optimisation algorithm. We use an image scale of 512x512 as the input for both training and testing. On URPC2017, the batch-size is 16, and the learning rate is 0.0001. Our models often diverge when we use a high learning rate due to unstable gradients, and all the detectors in the ensemble achieve the best performance after running 120 epochs. \textcolor{black}{On URPC2018 and URPC2019, the batch-size is 16. We first train each detector in the ensemble with a learning rate 0.001 for 80 epochs, and then train them with a learning rate 0.0001 for another 40 epochs. On ChinaMM, the batch-size is 16, and the learning rate is 0.001. Each detector in the ensemble runs 120 epochs.} The source code will be made available at:\url{https://github.com/LongChenCV/SWIPENET+CMA}.

\section{Ablation studies}
\label{sec:results}
In this section, we conduct the ablation experiments to investigate the influence of different components on our SWIPENET+CMA framework, \textcolor{black}{including the skip connection, the dilated convolution block and the CMA training paradigm. In the next section, we compare our method against several state-of-the-art (SOAT) detection frameworks on four datasets}.
\begin{table}[htbp]
\begin{center}
\caption{\textcolor{black}{Ablation studies on four datasets. Skip indicates skip connection, and Dilation indicates dilated convolution block. mAP indicates mean Average Precision(\%).}}
\begin{tabular}{c|c|c|c|c}
\hline
Dataset & Network & Skip & Dilation & mAP\\
\hline
\multirow{3}{*}{URPC2017} & UWNET1 & & \checkmark & 40.4\\
& UWNET2 & & & 38.3\\
& SWIPENET & \checkmark & \checkmark & 42.1\\
\hline
\multirow{3}{*}{URPC2018} & UWNET1 & & \checkmark & 61.2\\
& UWNET2 & & & 58.1\\
& SWIPENET & \checkmark & \checkmark & 62.2\\
\hline
\multirow{3}{*}{URPC2019} & UWNET1 & & \checkmark & 55.0\\
& UWNET2 & & & 54.2\\
& SWIPENET & \checkmark & \checkmark & 57.6\\
\hline
\multirow{3}{*}{ChinaMM} & UWNET1 & & \checkmark & 73.9\\
& UWNET2 & & &  71.0\\
& SWIPENET & \checkmark & \checkmark & 76.1\\
\hline
\end{tabular}
\end{center}
\label{tab:abla1}
\end{table}
\begin{table*}[htbp]
\caption{The performance (mAP(\%)) of SWIPENET in each iteration of CMA on test set of four datasets. The red numbers indicate the results of the 'clean' SWIPENETs.}
\begin{center}
\begin{tabular}{cc|ccccc|ccccccc}
\hline
\multirow{2}{*}{Dataset} & Stage & \multicolumn{5}{c|}{NECMA} & \multicolumn{7}{c}{NLCMA}\\
\cline{2-14}
& Iteration & 1 & 2 & 3 & 4 & 5 & 1 & 2 & 3 & 4 & 5 & 6 & 7\\
\hline
\multirow{2}{0.5in}{URPC2017} & Single & 42.1 & 44.2 & \textcolor{red}{45.3} & 40.5 & 37.2 & 47.5 & 47.2 & 46.2 & 47.9 & \textbf{48.0} & 47.0 & 47.6\\
& Ensemble & 42.1 & 45.0 & \textcolor{red}{46.3} & 45.3 & 44.2 & 47.5 & 48.6 & 49.8 & 52.3 & \textbf{52.5} & 52.5 & 52.5\\
\hline
\multirow{2}{0.5in}{URPC2018} & Single & 62.2 & \textcolor{red}{63.3} & 62.4 & 61.2 & 59.3 & 65.0 & 64.8 & \textbf{65.3} & 64.5 & 64.5 & 63.9 & 64.3\\
& Ensemble & 62.2 & \textcolor{red}{64.5} & 64.0 & 62.8 & 62.1 & 65.0 & 65.4 & 66.9 & 67.5 & \textbf{68.0} & 68.0 & 68.0\\
\hline
\multirow{2}{0.5in}{URPC2019} & Single & 57.6 & \textcolor{red}{58.5} & 57.2 & 56.9 & 56.1 & \textbf{61.8} & 61.5 & 61.6 & 61.0 & 59.5 & 61.5 & 61.0\\
& Ensemble & 57.6 & \textcolor{red}{59.9} & 59.0 & 59.0 & 59.5 & 61.8 & 62.4 & \textbf{63.9} & 63.9 & 63.9 & 63.9 & 63.9\\
\hline
\multirow{2}{0.5in}{ChinaMM} & Single & 76.1 & 77.5 & \textcolor{red}{78.3} & 76.5 & 74.8 & 80.4 & 79.8 & \textbf{82.3} & 81.4 & 79.5 & 80.0 & 79.3\\
& Ensemble & 76.1 & 78.5 & \textcolor{red}{79.9} & 77.8 & 78.5 & 80.4 & 81.9 & 83.4 & \textbf{85.6} & 85.5 & 85.6 & 85.6\\
\hline
\end{tabular}
\end{center}
\label{tab:CMA2017}
\end{table*}
\begin{figure*}[tbp]
\centerline{\includegraphics[width=18cm, height=5cm]{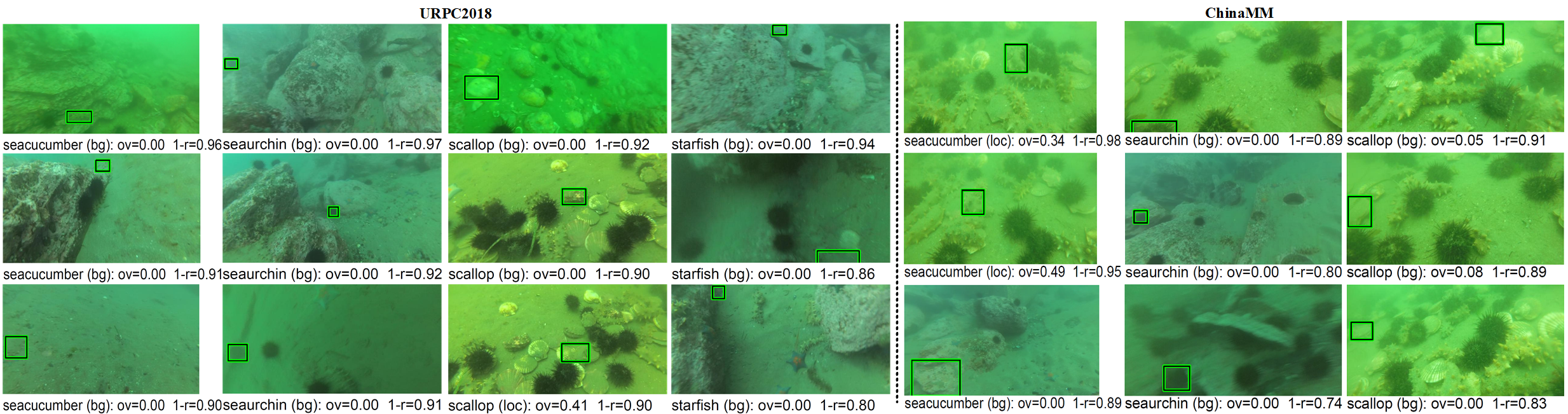}}
\textcolor{black}{\caption{Examples of top false positives of the SWIPENET without CMA: We show the top three false positives (FPs) for all categories on URPC2018 and ChinaMM. The text indicates the type of error ("loc"=localization; "bg"=confusion with backgrounds), the amount of overlap ("ov") with a true object, and the fraction of correct examples that are ranked lower than the given false positive ("1-r", for 1-recall). Localization errors are due to insufficient overlaps.}}
\label{fig:tpfp}
\end{figure*}
\subsection{Ablation studies on the skip connection and dilated convolution block}
\label{abls}
To investigate the influence of skip connection, we design the first baseline network UWNET1 which has the same structure as SWIPENET except that it does not contain skip connection between the low and high layers. The second network UWNET2 replaces the dilated convolution block in UWNET1 with standard convolution block to learn the influence of the dilated convolution block. \textcolor{black}{Table~\ref{tab:abla1} shows the performance comparison of different networks on four datasets, we observe that SWIPENET performs better than UWNET1. The gains come from the skip connection which passes fine detailed information of the lower layers such as object boundary to the high layers that are important for object localisation. Compared to UWNET2, UWNET1 performs better because the dilated convolution block in UWNET1 brings much semantic information to the high layers which enhances the classification ability. We also present the mean Average Precision (mAP) of UWNET2 and SWIPENET for the objects with different sizes in Fig.~\ref{fig:ObjectSize} and Supplementary Fig. 2, from which we observe the skip connection and dilated convolution block largely improves the small object detection accuracy. For example, for small objects (S) of seacucumber, seaurchin and scallop categories, SWIPENET improves 5\%$\sim $6\% mAP over UWNET2 on URPC2018 and ChinaMM.}
\begin{figure*}[tbp]
\centerline{\includegraphics[width=18cm, height=6cm]{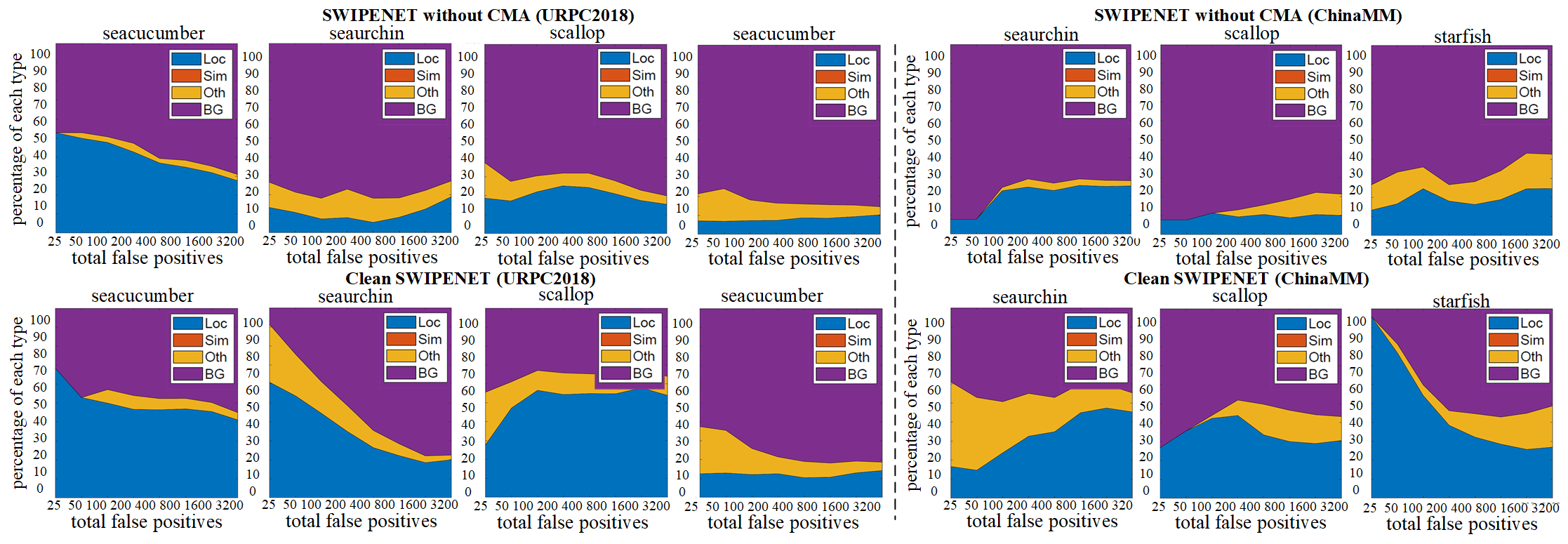}}
\textcolor{black}{\caption{The distribution of top-ranked false positive types of the SWIPENET without CMA and the 'clean' SWIPENET for each category on URPC2018 and ChinaMM. The false positive types include localisation error (Loc), confusion with similar categories (Sim), with others (Oth), or with background (BG).}}
\label{fig:errors}
\end{figure*}
\begin{figure*}[tbp]
\centerline{\includegraphics[width=18cm, height=5cm]{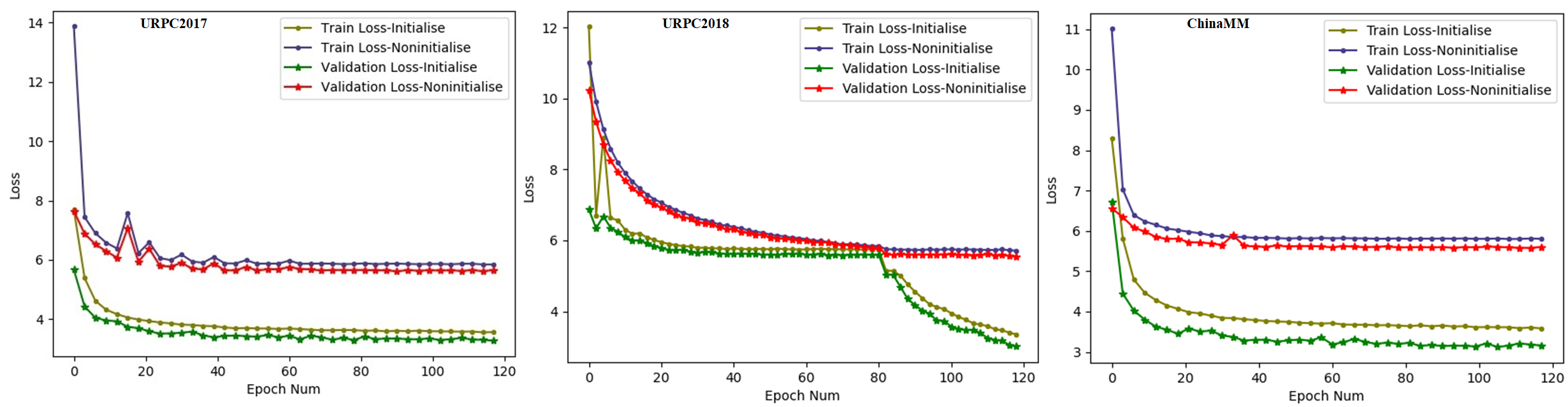}}
\textcolor{black}{\caption{The learning curve of SWIPENETs with and without initialisation by the 'clean' SWIPENET.}}
\label{fig:loss}
\end{figure*}

\subsection{Ablation studies on CMA}
In this subsection, we investigate the influence of CMA, including NECMA and NLCMA, on the final detection results. In our experiments, the number of the iterations of NECMA is set to 5 and the number of the iterations of NLCMA is set to 7. \textcolor{black}{Table \ref{tab:CMA2017} shows the performance of the single model and the ensemble model after each iteration on the testing set of four datasets.}\\
\textcolor{black}{\textbf{The role of NECMA.} From Table~\ref{tab:CMA2017}, in the noise-eliminating stage (NECMA), we observe that the 'clean' SWIPENET is achieved in the 3rd iteration on URPC2017 and ChinaMM, and in the 2nd iteration on URPC2018 and URPC2019. So we set $M_1=3$ on UPRC2017 and ChinaMM, and $M_1=2$ on URPC2018 and URPC2019. The 'clean' SWIPENETs perform much better than the detectors in the 1st iteration. We assume this is because the noisy data (backgrounds but labelled as objects categories in the annotation process) confuse the detectors in the 1st iteration. Fig.~\ref{fig:tpfp} and Supplementary Fig. 3 show the top three false positives for the 1st detector, i.e. the SWIPENET trained without CMA, we can see that the background error (detecting the backgrounds as the objects) has much influence on the detectors than the localisation error (inaccurate localisation). To further verify this assumption, we use the detection analysis tool of \cite{b44} to analyse the false positives of the 1st detector and the 'clean' detector in NECMA. Fig.~\ref{fig:errors} and Supplementary Fig. 4 show the distribution of the top-ranked false positives for each category on four datasets. We can see that the 1st detector cannot well distinguish the objects with complex background and generate much more background errors than the 'clean' detector. NECMA gradually reduces the influence of the noisy data on the single detector by decreasing their weights, and the background error clearly decreases in the detection results of the 'clean' SWIPENET. However, the performance of the single detectors after the 'clean' SWIPENET is less satisfactory. This is because most of the detected objects are continuously up-weighted and the detectors over-fit over these high-weight objects.}
 
\textbf{The role of NLCMA.} In the noise-learning stage (NLCMA), we initialise each detector using the parameter learned in the 'clean' SWIPENET. This strategy provides a good initialisation for the following detectors which avoid getting stuck in poor local minima during the training. With this initialisation strategy, the detectors converge much faster during the training, shown in Fig.~\ref{fig:loss} (we also take the testing set as the validation set and investigate the influence of this initialisation strategy on the validation loss). \textcolor{black}{From Table \ref{tab:CMA2017}, we can see all the detectors in NLCMA perform better than the 'clean' detectors. This is because the detectors in NECMA take all the undetected objects as the noisy data and ignore learning them, however, in addition to the noisy data, the undetected objects also contain many hard targets, which are hard to be detected due to their minor discrepancies with the backgrounds. The 'clean' detector trained by NECMA can only detect the easy objects well but mis-detect many hard targets that limits the generalization of the detector. Different from detectors in NECMA, the detectors in NLCMA are able to detect the hard targets with the help of the 'clean' SWIPENET. The fundamental knowledge learnt by the 'clean' SWIPENET helps the following detectors identify the minor discrepancies between the hard targets and the backgrounds.}

\begin{table*}[htbp]\scriptsize
\textcolor{black}{\caption{Comparison with small object detection frameworks on URPC2017, URPC2018, and ChinaMM.}}
\begin{center}
\setlength{\tabcolsep}{1mm}{
\begin{tabular}{l|cccc|ccccc|cccc}
\hline
Dataset & \multicolumn{4}{c|}{URPC2017} & \multicolumn{5}{c}{URPC2018} & \multicolumn{4}{c}{ChinaMM}\\
\hline
Methods & seacucumber & seaurchin & scallop & mAP & seacucumber & seaurchin & scallop & starfish & mAP & seacucumber & seaurchin & scallop & mAP\\
\hline
DSSD \cite{b33} & 13.2 & 70.3 & 26.4 & 36.6 & 48.4 & 75.3 & 38.2 & 64.0 & 56.5 & 54.5 & 82.0 & 79.4 & 72.0\\
FCOS \cite{b48} & 23.6 & \textbf{75.2} & 33.8 & 44.2 & 43.2 & 76.5 & 47.5 & 69.4 & 59.1 & 57.7 & 83.1 & 78.7 & 73.2\\
RetinaNet \cite{b26} & 19.2 & 72.9 & 28.9 & 40.3 & 52.5 & 74.9 & 43.1 & 69.8 & 60.1 & 59.6 & 82.0 & 81.0 & 74.2\\
FRCNN-FPN \cite{b52} & 25.3 & 73.7 & 26.3 & 41.8 & 57.7 & 76.9 & 38.1 & 70.6 & 60.9 & 58.0 & 82.1 & 81.6 & 73.9\\
RetinaNet with S-$\alpha$ \cite{b53} & 27.2 & 74.6 & 31.6 & 44.5 & 54.4 & 76.5 & \textbf{52.4} & 71.7 & 63.8 & 60.8 & 82.0 & 82.7 & 75.2\\
FRCNN-FPN with S-$\alpha$ \cite{b53} & 18.7 & 75.0 & 39.6 & 44.4 & \textbf{59.1} & 77.0 & 39.2 & 71.4 & 61.7 & 62.0 & 82.4 & 82.7 & 75.7\\
SWIPENET-noCMA & 43.6 & 51.3 & 31.2 & 42.1 & 46.4 & \textbf{84.0} & 40.2 & 78.2 & 62.2 & 63.0 & 83.5 & 81.9 & 76.1\\
SWIPENET-Single & \textbf{46.6} & 55.8 & \textbf{41.6} & \textbf{48.0} & 54.8 & 81.5 & 46.6 & \textbf{78.4} & \textbf{65.3} & \textbf{77.0} & \textbf{84.7} & \textbf{85.2} & \textbf{82.3}\\
\hline
\end{tabular}}
\end{center}
\label{tab:sotsmall}
\end{table*}

\begin{table*}[htbp]\scriptsize
\textcolor{black}{\caption{Comparison with underwater object detection frameworks on URPC2017, URPC2018, and ChinaMM.}}
\begin{center}
\setlength{\tabcolsep}{1mm}{
\begin{tabular}{ll|cccc|ccccc|cccc}
\hline
\multicolumn{2}{c|}{Dataset} & \multicolumn{4}{c|}{URPC2017} & \multicolumn{5}{c}{URPC2018} & \multicolumn{4}{c}{ChinaMM}\\
\hline
Methods & Backbone & seacucumber & seaurchin & scallop & mAP & seacucumber & seaurchin & scallop & starfish & mAP & seacucumber & seaurchin & scallop & mAP\\
\hline
SSD \cite{b8} & VGG16 \cite{b36} & 38.4 & 52.9 & 15.7 & 35.7 & 44.2 & 84.4 & 35.8 & 78.1 & 60.6 & 47.3 & 80.3 & 78.1 & 68.6\\
YOLOv3 \cite{b43} & DarkNet53 \cite{b43} & 28.4 & 50.3 & 22.4 & 33.7 & 35.7 & 83.0 & 34.0 & 77.9 & 57.7 & 33.1 & 80.2 & 77.9 & 63.7\\
FRCNN \cite{b20} & VGG16 \cite{b36} & 27.2 & 45.0 & 31.9 & 34.7 & 43.3 & 83.0 & 32.0 & 74.5 & 58.2 & 38.5 & 77.9 & 77.1 & 64.5\\
FRCNN \cite{b20} & ResNet50 \cite{b45} & 31.0 & 41.4 & 33.5 & 35.3 & 41.1 & 83.2 & 34.5 & 77.2 & 59.0 & 41.0 & 81.0 & 78.1 & 66.7\\
FRCNN \cite{b20} & ResNet101 \cite{b45} & 26.2 & 47.7 & 32.5 & 35.5 & 44.3 & 82.5 & 34.7 & 77.5 & 59.8 & 51.7 & 81.5 & 79.5 & 70.9\\
FRCNN \cite{b20} & FPN \cite{b52} & 25.3 & 73.7 & 26.3 & 41.8 & 57.7 & 76.9 & 38.1 & 70.6 & 60.9 & 58.0 & 82.1 & 81.6 & 73.9\\
IMA \cite{b42} & SWIPENET \cite{b42} & 44.4 & 52.4 & 42.1 & 46.3 & 52.8 & 84.1 & 42.9 & 78.0 & 64.5 & 68.3 & 83.3 & 84.5 & 78.7\\
RetinaNet \cite{b26} & ResNet50 \cite{b45} & 19.2 & 72.9 & 28.9 & 40.3 & 52.5 & 74.9 & 43.1 & 69.8 & 60.1 & 59.6 & 82.0 & 81.0 & 74.2 \\
FCOS \cite{b48} & ResNet50 \cite{b45} & 23.6 & 75.2 & 33.8 & 44.2 & 43.2 & 76.5 & 47.5 & 69.4 & 59.1 & 57.7 & 83.1 & 78.7 & 73.2\\
FreeAnchor \cite{b49} & ResNet50 \cite{b45} & 21.8 & 74.7 & 27.7 & 41.4 & 46.2 & 72.3 & 42.5 & 71.4 & 58.1 & 41.9 & 80.6 & 76.9 & 66.4\\ 
GHM \cite{b50} & ResNet50 \cite{b45} & 23.0 & 74.3 & 33.9 & 43.7 & 52.4 & 78.4 & 42.1 & 71.5 & 61.1 & 53.7 & 82.1 & 82.3 & 72.7\\
\hline
SWIPENET-Single & SWIPENET & 46.6 & 55.8 & 41.6 & 48.0 & 54.8 & 81.5 & 46.6 & 78.4 & 65.3 & 77.0 & 84.7 & 85.2 & 82.3\\
SWIPENET-CMA & SWIPENET & \textbf{49.1} & \textbf{62.3} & \textbf{46.1} & \textbf{52.5} & \textbf{56.4} & \textbf{84.6} & \textbf{50.9} & \textbf{79.9} & \textbf{68.0} & \textbf{82.2} & \textbf{87.1} & \textbf{87.6} & \textbf{85.6}\\
\hline
\end{tabular}}
\end{center}
\label{tab:sotuod}
\end{table*}

\subsection{Ablation studies on the selective ensemble algorithm.}
\begin{figure}[tbp]
\centerline{\includegraphics[width=7.5cm, height=6cm]{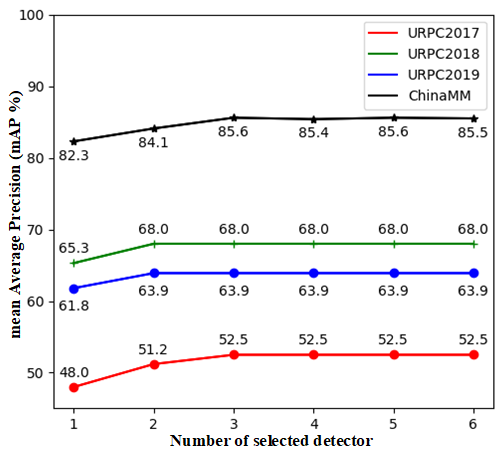}}
\textcolor{black}{\caption{The performance of the ensemble with different numbers of detectors.}}
\label{fig:mAP_Num}
\end{figure}
We investigate the influence of selective ensemble algorithm (SE) on the performance of the final ensemble detector. \textcolor{black}{Fig.~\ref{fig:mAP_Num} shows the performance of the ensemble detector with different numbers of the selected detectors. The SE algorithm reduces the number of the detectors in the final ensemble. For example, the ensemble detector without SE achieves the best mAP on URPC2017 and URPC2018 when we ensemble five detectors, but the ensemble detector with SE achieves the same mAP by only integrating three selected detectors on URPC2017 and two selected detectors on URPC2018. This demonstrates some of the detectors do not help boosting the final performance in the ensemble. Few detectors with large diversity are sufficient to achieve the best performance. The selective ensemble algorithm surely helps reduce the computational overhead during testing due to the reduced number of the detectors.}

\begin{figure*}[htbp]
\centerline{\includegraphics[width=18cm, height=6cm]{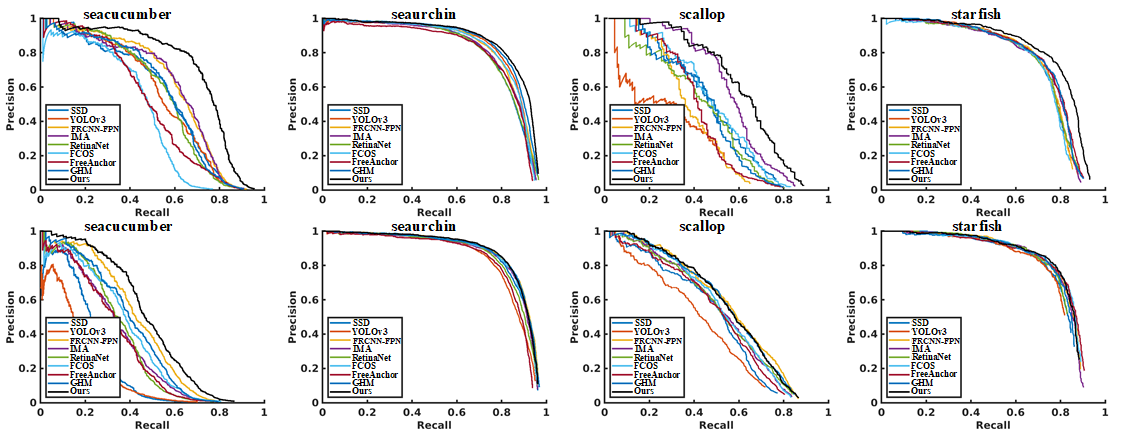}}
\caption{Precision/Recall curves of different detection methods on URPC2018 (top row) and URPC2019 (bottom row).}
\label{fig:roc18}
\end{figure*}

\begin{figure*}[htbp]
\centerline{\includegraphics[width=18cm, height=10cm]{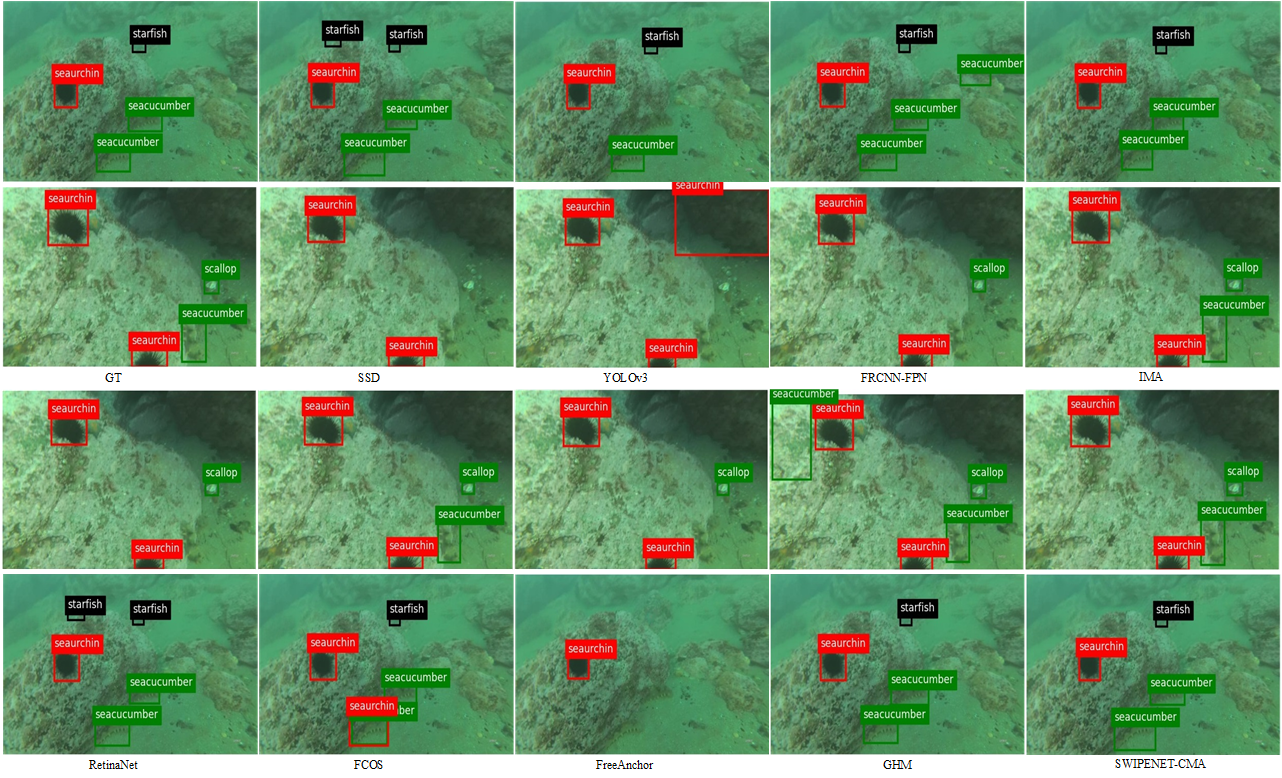}}
\caption{Visualization of object detection results of different detection frameworks. From left to right are raw underwater images with ground truth, results of SSD \cite{b8}, YOLOv3 \cite{b43}, FRCNN-FPN \cite{b52}, IMA \cite{b42}, RetinaNet \cite{b26}, FCOS \cite{b48}, FreeAnchor \cite{b49}, GHM \cite{b50} and our SWIPENET-CMA. The images in the top row come from URPC2018, and the images in the bottom row come from ChinaMM.}
\label{fig:visualisationURPC2018}
\end{figure*}

\section{Comparison with SOAT detection frameworks}
\label{contrast}
\textcolor{black}{In this section, we compare our proposed deep detector with other state-of-the-art (SOAT) detection frameworks. Since our SWIPENET are specially designed to improve small object detection, we first compare our proposed method with latest small object detection methods. Then, we compare SWIPENET+CMA framework with underwater object detection frameworks used in recent literatures.}

\textcolor{black}{\subsection{Comparison with small object detection frameworks}}\label{contrast}
\textcolor{black}{Following the latest small object detection work \cite{b47}, we select DSSD \cite{b33}, RetinaNet \cite{b26}, FCOS \cite{b48}, FRCNN-FPN \cite{b52}, and layer fusion strategy S-alpha \cite{b53} as the small object detection comparison methods. For fair comparison, we only compare our single models SWIPENET-noCMA (the SWIPENET trained without CMA) and SWIPENET-Single (the best single model achieved in the CMA) with other detection frameworks without considering the ensemble model.}

\textcolor{black}{\textbf{Implementation details.} For RetinaNet and FCOS, we use ResNet50 \cite{b45} as the backbone network. For DSSD and FRCNN-FPN, we use their original backbone networks. Following \cite{b53}, we use FRCNN-FPN and RetinaNet with layer fusion strategy S-alpha as the detection frameworks. Both use ResNet50 \cite{b45} backbone. The comparison methods are tuned to have the best performance.}

\textcolor{black}{The experimental results on URPC2017, URPC2018 and ChinaMM are shown in Table~\ref{tab:sotsmall}, from which we obeserve SWIPENET-noCMA performs much better than DSSD, this is because multiple down-sampling operations lost many useful features, which are importance for accurate small object localization, these features cannot fully recovered by up-sampling operations once lost. The dilated convolution block in SWIPENET retains these features that benefits object localisation. On three datasets, our SWIPENET-Single achieves the best performance, its advantage comes from the SWIPENET backbone and the noisy eliminating strategy. It is worth noting that FCOS and RetinaNet and FRCNN-FPN frameworks apply much deeper backbones (ResNet50) than our SWIPENET, but SWIPENET-noCMA still achieves better performance than the former three frameworks on URPC2018 and ChinaMM, this demonstrates the multiple Hyper Features in SWIPENET is able to detect multi-scale objects well. FRCNN-FPN with S-alpha achieves the best performance on URPC2019 (the results can be found in Supplementary TABLE I), this is because the layer fusion strategy S-alpha greatly boost the performance of small object detection, but it cannot solve the noise problem.}

\subsection{Comparison with underwater object detection frameworks}
\label{contrast}
\textcolor{black}{We also compare our method against several detection frameworks have ever applied for underwater object detection in recent literatures \cite{b42, b51}, we only select the comparision methods whose source code is public availiable online, including IMA \cite{b42}, SSD \cite{b8}, YOLOv3 \cite{b43}, FRCNN \cite{b20}, RetinaNet \cite{b26}, FCOS \cite{b48}, FreeAnchor \cite{b49} and GHM \cite{b50}.}

\begin{figure*}[htbp]
\centerline{\includegraphics[width=18cm, height=4.5cm]{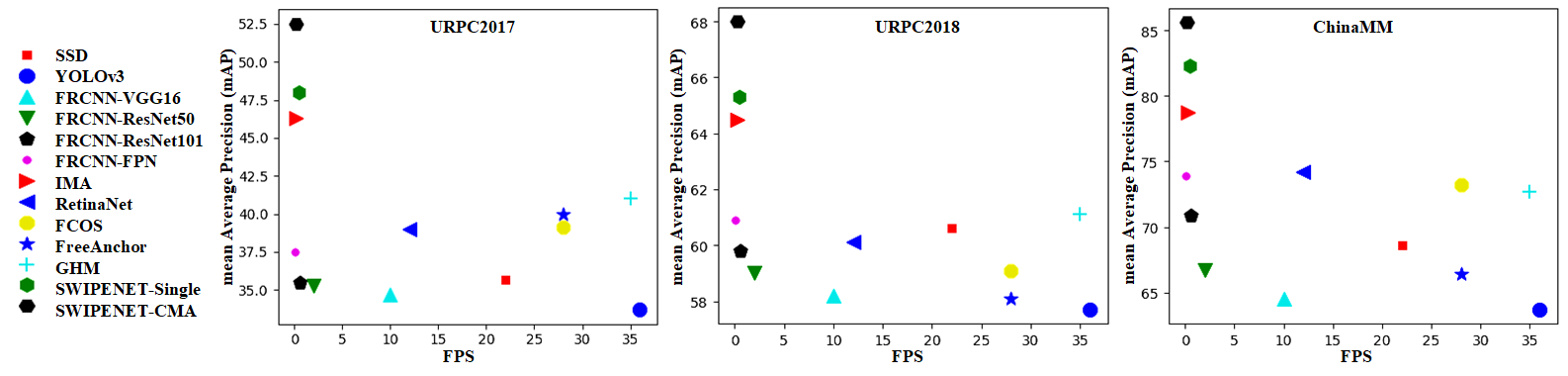}}
\textcolor{black}{\caption{Running time (Frames Per Second, FPS) vs mean Average Precision (mAP) of different detection frameworks.}}
\label{fig:FPS_mAP}
\end{figure*}

\textbf{Implementation details.} \textcolor{black}{For SSD, we use VGG16 \cite{b36} as the backbone. For Faster RCNN, we use four backbones including VGG16, ResNet50 \cite{b45}, ResNet101 \cite{b45} and FPN \cite{b52}. For YOLOv3, we use its original DarkNet53 network. RetinaNet, FCOS, FreeAnchor and GHM all use ResNet50 \cite{b45} as the backbones. The comparison methods are tuned to have the best performance.}

\textcolor{black}{Tables \ref{tab:sotuod} shows the experimental results on URPC2017, URPC2018 and ChinaMM, where our proposed SWIPENET-CMA achieves the best performance than other comparison methods. On three datasets, FRCNN with FPN performs better than FRCNN with ResNet101, ResNet50 and VGG16, where the deeper backbone FPN plays a critical role. SWIPENET-Single, the best single SWIPENET trained using CMA, outperforms the other frameworks by a large margin on three datasets, demonstrating the superiority of our proposed CMA in dealing with noisy data. It performs even better than the ensemble model trained with the IMA algorithm. This is because IMA regards all the undetected objects as noisy data and ignore learning them, which loses considerable effective training samples. Although the undetected objects tend to be noisy data or outliers, they also contain many hard object instances. Ignoring these hard object instances, IMA can only detect the easy objects well but cannot detect many hard objects. Similarly, GHM avoids learning both noisy data and hard objects, it can avoid the influence of the noisy data but cannot generalize well on the hard object instances. RetinaNet is easily to overfit on the noisy data because it employed the focal loss to train the detection network which emphasis on learning hard samples and also noisy data. Different from IMA and GHM, NLCMA stage of CMA focuses on learning possible hard object instances by increasing their weights, that improve the generalization on hard objects instances. SWIPENET-CMA further improves the SWIPENET-Single. The gain comes from its capacity to detect the diverse hard object instances. Fig.~\ref{fig:roc18} and Supplementary Fig. 7 show the Precision/Recall curves of different detection methods on four datasets, where we observe SWIPENET-CMA (black curve) achieves the best performance across all the object categories on URPC2017, URPC2018 and ChinaMM. Fig. \ref{fig:visualisationURPC2018} present visualization of object detection results of different detection frameworks on URPC2018 and ChinaMM (the visualization on URPC2017 and URPC2019 can be found in the Supplementary Fig. 6 and Fig. 7), most of the detection frameworks cannot detect all the objects, some of them detected the backgrounds as the objects. Among them, SWIPENET-CMA performs best.}

\textcolor{black}{On URPC2019, as shown in Supplementary TABLE I, our proposed method ranks the 2nd place in all comparison methods. The best performance is achieved by the method FRCNN-FPN with S-alpha, which is specially designed for small object detection. We think two factors explain why FRCNN-FPN with S-alpha performs better than our proposed method on URPC2019: First, FRCNN-FPN with S-alpha applied the much deeper backbone Feature Pyramid Networks (FPN) and advanced layer fusion strategy S-alpha to boost the performance of small object detection. Second, URPC2019 has more accurate annotations (less noisy labels) than URPC2017 and URPC2018. The challenge from small objects is much bigger than that from the noisy data on URPC2019. With less influence of the noisy data, FRCNN-FPN with S-alpha performs better than our proposed method.}

\begin{table*}[htbp]
\caption{The performance (mAP(\%)) of SWIPENET in each iteration of different training paradigm on the test set of URPC2017, URPC2018 and ChinaMM.}
\begin{center}
\begin{tabular}{l|l|ccccccccc}
\hline
Dataset & Iteration & 1 & 2 & 3 & 4 & 5 & 6 & 7 & 8\\
\hline
\multirow{4}{0.5in}{URPC2017} & SWIPENET+CMA & 42.1 & 45.0 & 46.3 & 47.5 & 48.6 & 49.8 & 52.3 & \textbf{52.5}\\
& SWIPENET+MA & \textbf{42.1} & 41.0 & 40.5 & 39.2 & 39.5 & 38.8 & 40.2 & 39.8\\
& SWIPENET+Curriculum &42.1 & 41.0 & \textbf{43.9} & - & - & - & - & -\\
\hline
\multirow{4}{0.5in}{URPC2018} & SWIPENET+CMA & 62.2 & 64.5 & 65.0 & 65.4 & 66.9 & 67.5 & \textbf{68.0} & 68.0\\
& SWIPENET+MA & \textbf{62.2} & 62.0 & 61.0 & 61.2 & 60.1 & 58.8 & 60.2 & 59.3\\
& SWIPENET+Curriculum &62.2 & 62.1 & \textbf{63.8} & -& - & - & - & -\\
\hline
\multirow{4}{0.5in}{URPC2019} & SWIPENET+CMA & 57.6 & 59.9 & 61.8 & 62.4 & \textbf{63.9} & 63.9 & 63.9 & 63.9\\
& SWIPENET+MA & \textbf{57.6} & 56.2 & 57.0 & 57.6 & 56.9 & 56.8 & 55.8 & 56.3\\
& SWIPENET+Curriculum & 57.6 & 56.9 & \textbf{60.8} & -& - & - & - & -\\
\hline
\multirow{4}{0.5in}{ChinaMM} & SWIPENET+CMA & 76.1 & 78.5 & 79.9 & 80.4 & 81.9 &83.4 & \textbf{85.6} & 85.5\\
& SWIPENET+MA & 76.1 & \textbf{77.0} & 76.5 & 76.0 & 75.5 & 75.7 & 75.0 & 74.7\\
& SWIPENET+Curriculum & 76.1 & 75.5 & \textbf{78.2} & - & - & - & - & -\\
\hline
\end{tabular}
\end{center}
\label{tab:Paradigm}
\end{table*}

\subsection{Comparison with representative learning paradigms}
\label{paradigm}
\textcolor{black}{CMA combines the learning tricks from Multi-Class Adaboost \cite{b42} and Curriculum Learning \cite{b32}, hence, we also conduct additional experiments to further compare our CMA learning paradigm with these two learning paradigms.}\\
\textcolor{black}{\textbf{Implementation details.} \textbf{SWIPENET+MA} train multiple detectors using the Multi-Class Adaboost algoithm and finally ensemble them into a unified model, focusing on learning undetected samples by up-weighting their weights. \textbf{SWIPENET+Curriculum} first trains a detector on the easy samples, then fine-tunes the detector of hard samples, since curriculum paradigm needs to define the easy and hard training samples: Similar to \cite{b12} that takes misclassified samples as the hard samples, we take the undetected objects as hard samples and the detected objects as easy samples. Specially, we first train a detector on all the training data, then we test the detector on the training data, the detected objects as easy and undetected objects as hard samples.}

\textcolor{black}{Table~\ref{tab:Paradigm} shows the performance comparison of different training paradigms. Our CMA performs much better than the other training paradigms on all four datasets. After the 1st iteration, MA enable the detectors to focus on learning the hard data that degrade the system performance. This is because these hard data contain many noisy data confuse the detectors. On the four datasets, Curriculum decays the performance in the 2nd iteration but boosts the performance in the 3rd iteration. This is because Curriculum trains the detector using insufficient easy samples in the 2nd iteration. After having fine-tuned over the remaining hard samples, the performance is better than that in the 1st iteration. The gains come from the easy-to-hard training strategy and sufficient training data. However, CMA still performs much better than Curriculum. This is because the underwater datasets contain considerable diverse data resources due to frequently changing illuminations and environments, the ensemble model is able to learn diverse data and performs much better than the single model trained using the Curriculum paradigm whose generalisation ability is limited.}

\section{Conclusion}
This paper offers a compelling insight on the training strategy of deep detectors in underwater scenes where noisy data exist. We have presented a new neural network architecture, called Sample-WeIghted hyPEr Network (SWIPENET), for small underwater object detection. Moreover, a sample re-weighting algorithm named Curriculum Multi-Class Adaboost (IMA) had been presented to deal with the noise issue. \textcolor{black}{Our proposed method well-handles the noise issue in underwater object detection and achieves the state-of-the-art performance on the challenging underwater datasets. However, since it is an ensemble deep model, the time complexity is much higher than current popular single models (as shown in Fig. \ref{fig:FPS_mAP}). Hence, in our future work, reducing the computational complexity of our proposed method is of vital importance. In the future work, we will extend our proposed method to more general application sceneries where considerable noise exits.}
\section*{Acknowledgment}
Thanks for National Natural Science Foundation of China and Dalian Municipal People's Government providing the underwater object detection datasets for research purposes. Haiping Ma is supported by Zhejiang Provincial Natural Science Foundation of China under Grant No. LY19F030011.

\bibliographystyle{IEEEtran}
\bibliography{mybibfile}
\vspace{-12 mm}
\begin{IEEEbiographynophoto}{Long Chen} is currently pursuing the PhD degree with the School of Informatics, University of Leicester, U.K. His research interests are in the areas of Computer Vision and Machine Learning.
\end{IEEEbiographynophoto}
\vspace{-12 mm}
\begin{IEEEbiographynophoto}{Feixiang Zhou} is currently pursuing the Ph.D. degree with the School of Informatics, University of Leicester, Leicester, U.K.
\end{IEEEbiographynophoto}
\vspace{-12 mm}
\begin{IEEEbiographynophoto}{Shengke Wang} is currently a Associate Professor of the Department of Computer Science and Technology, Ocean University of China, Qingdao, China. 
\end{IEEEbiographynophoto}
\vspace{-12 mm}
\begin{IEEEbiographynophoto}{Junyu Dong} is currently a Professor and the Head of the Department of Computer Science and Technology, Ocean University of China, Qingdao, China.
\end{IEEEbiographynophoto}
\vspace{-12 mm}
\begin{IEEEbiographynophoto}{Ning Li} currently is an Associate Professor at College of Electronic and Information Engineering, Nanjing University of Aeronautics and Astronautics, China.
\end{IEEEbiographynophoto}
\vspace{-12 mm}
\begin{IEEEbiographynophoto}{Haiping Ma} is an Associate Professor at Department of Electrical Engineering, Shaoxing University, Shaoxing, Zhejiang, 312000, China.
\end{IEEEbiographynophoto}
\vspace{-12 mm}
\begin{IEEEbiographynophoto}{Xin Wang} is an Associate Professor at College of Computer and Information, Hohai University, China.
\end{IEEEbiographynophoto}
\vspace{-12 mm}
\begin{IEEEbiographynophoto}{Huiyu Zhou} currently is a Professor at School of Informatics, University of Leicester, United Kingdom.
\end{IEEEbiographynophoto}

\newpage
\onecolumn
\section*{Supplementary}
\textcolor{black}{\subsection{The implementation of the dilated convolution block}}
The proposed dilated convolution block consists of 4 dilated convolution layers with ReLU activation. Denote the input and output of $i$-th dilated convolution layer as $F_I^i$ and $F_O^i$. The input of dilated block is the feature maps from the Pool5 layer ($F_{P5}$) and the output of dilated block $F_O^4$ is implemented using the following steps: Denote the $i$-th convolution kernel as $\theta_i\in \mathbb{R}^{K_i\times K_i\times C_i}$ ($K_i$ is the kernel size and $C_i$ is the channel number). Different from convolution operation (denote as $*$), dilated convolution operation first produce a dilated kernel $\theta_i^d$ by dilating the convolution kernel using the dilation function $\mathcal{F_D}(.)$ parametrized with a dilation rate $d_i$.
\begin{equation}
\theta_i^d=\mathcal{F_D}(\theta_i, d_i)\in \mathbb{R}^{K_i^d\times K_i^d\times C_i}
\label{formula:dilated}
\end{equation}
where 
\begin{equation}
K_i^d=K_i+(K_i-1)\times(d-1)
\label{formula:kernelsize}
\end{equation}
$\mathcal{F_D}(\theta_i, d_i)$ indicates insert $d_i$-1 zeros between neighbour values in each channel of $\theta_i$. Then, the dilated kernel is applied to convolution on the input. Finally, ReLU activation function $\mathcal{F_R}(.)$ is applied to the dilated convolution layer and produce the output of each dilated convolution layer.
\begin{equation}
F_O^i=\mathcal{F_R}(F_I^i*\mathcal{F_D}(\theta_i, d_i)), i=1,2,3,4
\label{formula:output}
\end{equation}
where 
\begin{equation}
F_I^i=\left\{\begin{matrix} F_{P5} \qquad\ \textbf{if } i=1
\\
F_O^{i-1} \quad \textbf{otherwise}
\end{matrix}\right.
\label{formula:outin}
\end{equation}

\textcolor{black}{\subsection{The detailed explanation of sample-weighted detection loss}}
\begin{figure}[h]
\centerline{\includegraphics[width=13cm, height=4cm]{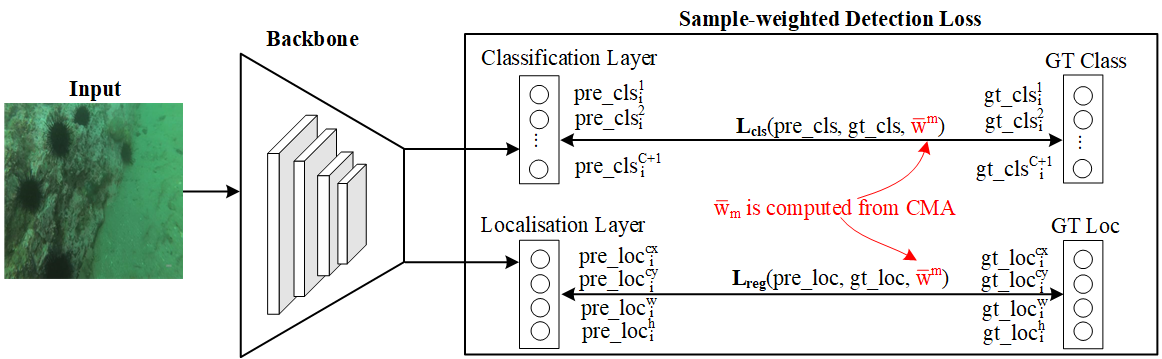}}
\caption{The detailed explanation of sample-weighted detection loss.}
\label{fig:DLoss}
\end{figure}
Fig. 1 presents the detailed explanation of sample-weighted detection loss. $pre\_cls_i$ and $gt\_cls_i$ denote the predicted and ground truth class vectors for the i-th sample respectively, these two vectors are C+1-D vectors (C object classes plus one background class). Here, $pre\_cls_i^c$ and $gt\_cls_i^c$ denote the c-th element of the predicted and ground truth class vectors for the i-th sample. $pre\_loc_i$ and $gt\_loc_i$ denote the predicted and ground truth coordinate vectors for the i-th sample, these two coordinate vectors are 4-D vectors, the coordinate vector Loc=(cx, cy, w, h) includes the coordinates of center (cx, cy) with width w and height h. Here, $pre\_loc_i^l$ and $gt\_loc_i^l$ denote the l-th element of the predicted and ground truth coordinate vectors for the i-th sample. $\bar{w}_m^i$ denotes the sample weight for the i-th sample computed in the m-th iteration of CMA.

\textcolor{black}{\subsection{The derivation the proposed sample-weighted detection loss}}

The proposed sample-weighted detection loss $L$ consists of a sample-weighted softmax loss $L_{cls}$ for the bounding box classification and a sample-weighted smooth L1 loss $L_{reg}$ for the bounding box regression:
\begin{equation}
L=\frac{\alpha_{1}}{\ddot{N}}L_{cls}(pre\_cls,gt\_cls)+\frac{\alpha_{2}}{\bar{N}} L_{reg}(pre\_loc,gt\_loc)
\label{formula:1}
\end{equation}
where the sample-weighted softmax loss $L_{cls}$ is formulated as follows:
\begin{equation}
\begin{split}
L_{cls}=-\sum_{i=1}^{\ddot{N}} \sum_{c=1}^{C+1} \bar{w}_{i}^{m}gt\_cls_{i}^{c}log(pre\_cls_{i}^{c})
\label{formula:2}
\end{split}
\end{equation}

\begin{equation}
pre\_cls_{i}^{c}=\frac{e^{net_i^c}}{\sum_{\bar c=1}^{C+1}e^{net_i^{\bar c}}}
\label{formula:3}
\end{equation}
and the sample-weighted smooth L1 loss $L_{reg}$  is formulated as follows:
\begin{equation}
L_{reg}=\sum_{i=1}^{\bar N}\sum_{l\in Loc}\bar{w}_{i}^{m}Smooth_{L_1}(pre\_loc_{i}^{l}-gt\_loc_{j}^{l})
\label{formula:reg}
\end{equation}
\begin{equation}
Smooth_{L_1}(x)=\left\{\begin{matrix} 0.5x^2 \qquad\ \textbf{if } |x|<1
\\
|x|-0.5 \quad \textbf{otherwise}
\end{matrix}\right.
\label{formula:smoothl1}
\end{equation}
\begin{equation}
pre\_loc_i^l=net_i^l, l \in Loc
\label{formula:smoothl1}
\end{equation}
As the gradient magnitude of the sample determines how much impact is applied to the updating of the DNNs, we investigate how the weight of the sample influences the feature learning of SWIPENET through the sample-weighted detection loss function. Let $\theta$ be the parameters of SWIPENET, and the gradient of $L$ with respect to $\theta$ can be derived as follows:
\begin{equation}
\begin{aligned}
\frac{\partial L}{\partial \theta} = \frac{\alpha_1}{\ddot{N}} \frac{\partial L_{cls}}{\partial \theta} + \frac{\alpha_2}{\bar N} \frac{\partial L_{reg}}{\partial \theta}
\end{aligned}
\end{equation}
To obtain $\frac{\partial L}{\partial \theta}$, we first derive $ \frac{\partial L_{cls}}{\partial \theta}$ as
\begin{equation}
\begin{aligned}
\frac{\partial L_{cls}}{\partial \theta} &=-\sum_{i=1}^{\ddot{N}} \sum_{c=1}^{C+1} \bar{w}_{i}^{m}gt\_cls_{i}^{c} \frac{\partial log(pre\_cls_{i}^{c})}{\partial \theta}\\
&=-\sum_{i=1}^{\ddot{N}} \sum_{c=1}^{C+1} \bar{w}_{i}^{m}gt\_cls_{i}^{c} \frac{1}{pre\_cls_{i}^{c}} \frac{\partial pre\_cls_i^c}{\partial net_i^c} \frac{\partial net_i^c}{\partial \theta}
\end{aligned}
\label{eq:Lclsderive}
\end{equation}
where
\begin{equation}
\begin{aligned}
\frac{\partial pre\_cls_i^c}{\partial net_i^c} &= \frac{e^{net_i^c}\sum_{c=1}^{C+1}e^{net_i^c}-(e^{net_i^c})^2}{(\sum_{c=1}^{C+1}e^{net_i^c})^2}\\
&=\frac{e^{net_i^c}}{\sum_{\bar c=1}^{C+1}e^{net_i^{\bar c}}}-(\frac{e^{net_i^c}}{\sum_{\bar c=1}^{C+1}e^{net_i^{\bar c}}})^2\\
&=pre\_cls_i^c-(pre\_cls_i^c)^2
\end{aligned}
\label{eq:precls}
\end{equation}
Substituting Eq.~(\ref{eq:precls}) into Eq.~(\ref{eq:Lclsderive}), then we have
\begin{equation}
\begin{aligned}
\frac{\partial L_{cls}}{\partial \theta} =\sum_{i=1}^{\ddot{N}} \sum_{c=1}^{C+1}\bar{w}_{i}^{m}gt\_cls_{i}^{c} (pre\_cls_i^c-1) \frac{\partial net_i^c}{\partial \theta}
\end{aligned}
\end{equation}
Secondly, we derive $\frac{\partial L_{reg}}{\partial \theta}$ as
\begin{equation}\small
\begin{aligned}
\frac{\partial L_{reg}}{\partial \theta} &=
\sum_{i=1}^{\bar N} \sum_{l \in Loc} \bar{w}_i^m \frac{\partial Smooth_{L_1}}{\partial pre\_loc_i^l} \frac{\partial pre\_loc_i^l}{\partial net_i^l} \frac{\partial net_i^l}{\partial \theta}\\
&=\left\{\begin{matrix} \sum_{i=1}^{\bar N} \sum_{l \in Loc} \bar{w}_{i}^{m}(pre\_loc_{i}^{l}-gt\_loc_{j}^{l}) \frac{\partial net_i^l}{\partial \theta}\\
\quad\quad\quad\quad\quad\quad\ \ \ \textbf{if } |pre\_loc_{i}^{l}-gt\_loc_{j}^{l}|<1
\\ \pm \sum_{i=1}^{\bar N} \sum_{l \in Loc} \bar{w}_{i}^{m} \frac{\partial net_i^l}{\partial \theta} \qquad \ \ \ \textbf{otherwise}
\end{matrix}\right.
\label{formula:smoothl1}
\end{aligned}
\end{equation}
Finally, we have
\begin{equation}\small
\begin{aligned}
\frac{\partial L}{\partial \theta}=&\left\{\begin{matrix} \frac{\alpha_1}{\ddot{N}}\sum_{i=1}^{\ddot{N}} \sum_{c=1}^{C+1}\bar{w}_{i}^{m}gt\_cls_{i}^{c} (pre\_cls_i^c-1) \frac{\partial net_i^c}{\partial \theta}\\
+ \frac{\alpha_2}{\bar N} \sum_{i=1}^{\bar N} \sum_{l \in Loc} \bar{w}_{i}^{m}(pre\_loc_{i}^{l}-gt\_loc_{j}^{l}) \frac{\partial net_i^l}{\partial \theta}\\
\quad\quad\quad\quad\quad\quad\quad\quad\quad\ \textbf{if } |pre\_loc_{i}^{l}-gt\_loc_{j}^{l}|<1
\\ \frac{\alpha_1}{\ddot{N}}\sum_{i=1}^{\ddot{N}} \sum_{c=1}^{C+1}\bar{w}_{i}^{m}gt\_cls_{i}^{c} (pre\_cls_i^c-1) \frac{\partial net_i^c}{\partial \theta}\\
\pm \frac{\alpha_2}{\bar N} \sum_{i=1}^{\bar N} \sum_{l \in Loc} \bar{w}_{i}^{m} \frac{\partial net_i^l}{\partial \theta} \qquad\qquad \textbf{otherwise}
\end{matrix}\right.\\
\end{aligned}
\label{eq:finalderive}
\end{equation}
\clearpage
\subsection{Description of the underwater robot picking contest}
\textcolor{black}{The underwater robot picking contest datasets were generated by National Natural Science Foundation of China and Dalian Municipal People's Government. The Chinese website is \url{http://www.cnurpc.org/index.html} and the English website is \url{http://en.cnurpc.org/}. The contest holds annually from 2017, consisting of online and offline object detection contests. In this paper, we use URPC2017, URPC2018 and URPC2019 datasets from the online object detection contest. To use the datasets, participants need to communicate with zhuming@dlut.edu.cn and sign a commitment letter for data usage: \url{http://www.cnurpc.org/a/js/2018/0914/102.html}.}

\subsection{More experimental results on URPC2017, URPC2019 and ChinaMM}
\noindent This section shows more experimental results on URPC2017, URPC2019 and ChinaMM. 

\begin{table}[h]\scriptsize
\textcolor{black}{\caption{Comparison with other detection frameworks on URPC2019.}}
\begin{center}
\setlength{\tabcolsep}{1mm}{
\begin{tabular}{l|ccccc}
\hline
Methods & seacucumber & seaurchin & scallop & starfish & mAP\\
\hline
DSSD & 22.8 & 79.8 & 43.2 & 75.3 & 55.3\\
FCOS  & 36.8 & 81.2 & 51.0 & 75.8 & 61.2\\
RetinaNet & 34.8 & 79.5 & 50.4 & 74.8 & 59.8\\
FRCNN-FPN  & 40.2 & 81.8 & 54.5 & 75.9 & 63.1\\ 
RetinaNet with S-alpha & 42.2 & 78.7 & 54.0 & 76.1 & 62.8\\
FRCNN-FPN with S-alpha & \textbf{46.3} & \textbf{82.8} & \textbf{54.9} & 75.7 & \textbf{64.9}\\
SWIPENET-noCMA & 28.9 & 79.9 & 48.3 & 73.3 & 57.6\\
SWIPENET-Single & 41.7 & 81.8 & 50.4 & 73.2 & 61.8\\
\hline                          
SSD & 24.3 & 80.1 & 46.4 & 74.3 & 56.3\\ 
YOLOv3 & 18.1 & 78.1 & 40.4 & 73.3 & 52.5\\
FRCNN-VGG16 & 20.9 & 79.1 & 43.5 & 73.2 & 54.2\\
FRCNN-ResNet50 & 22.8 & 79.8 & 43.2 & 75.3 & 55.3\\
FRCNN-ResNet101 & 25.4 & 79.1 & 46.4 & 74.6 & 56.4\\ 
FRCNN-FPN  & 40.2 & 81.8 & 54.5 & 75.9 & 63.1\\ 
IMA & 34.1 & 80.7 & 50.5 & 75.6 & 60.2\\
FreeAnchor & 32.7 & 73.7 & 48.1 & 75.5 & 57.5\\
GHM & 38.3 & 80.6 & \textbf{53.2} & 75.2 & 61.8\\
SWIPENET-Single & 41.7 & 81.8 & 50.4 & 73.2 & 61.8\\
SWIPENET-CMA & 44.8 & 81.8 & 53.0 & \textbf{76.1} & 63.9\\
\hline
\end{tabular}}
\end{center}
\label{tab:sot19}
\end{table}
Table I shows the quantitative results of different detection framework on URPC2019, where our proposed method ranks the 2nd place among all comparison methods. The best performance is achieved by the method FRCNN-FPN with S-alpha, which is specially designed for small object detection.

\begin{figure*}[h]
\centerline{\includegraphics[width=18cm, height=5.8cm]{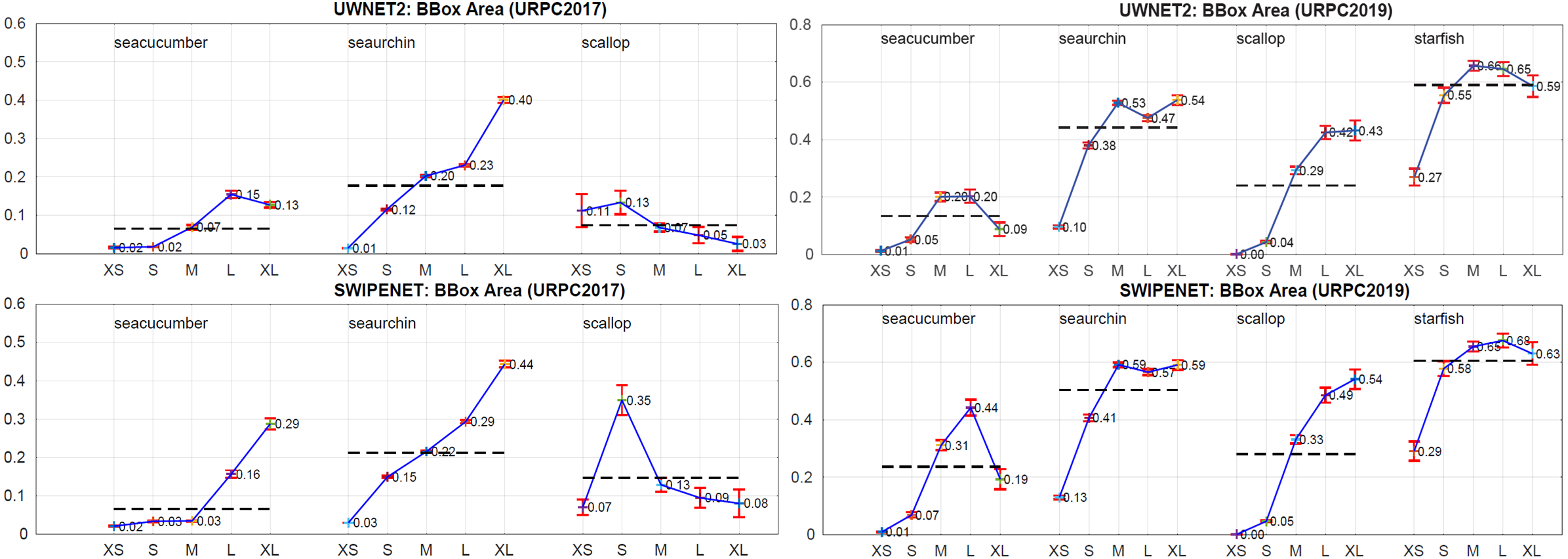}}
\caption{The mean Average Precision of UWNET2 and SWIPENET for objects with different object sizes on URPC2017 and URPC2019. The object size is measured as the pixel area of the bounding box. XS (bottom 10\%)=extra-small; S (next 20\%)=small; M (next 40\%)=medium; L (next 20\%)=large; XL (next 10\%)=extra-large.}
\label{fig:ObjectSize1719}
\end{figure*}
Fig.~\ref{fig:ObjectSize1719} presents the mAP of UWNET2 and SWIPENET for the objects with different sizes on URPC2017 and URPC2019. We observe that the skip connection and dilated convolution block improves the detection accuracy of objects with different scales.

\clearpage
\begin{figure*}[h]
\centerline{\includegraphics[width=18cm, height=5cm]{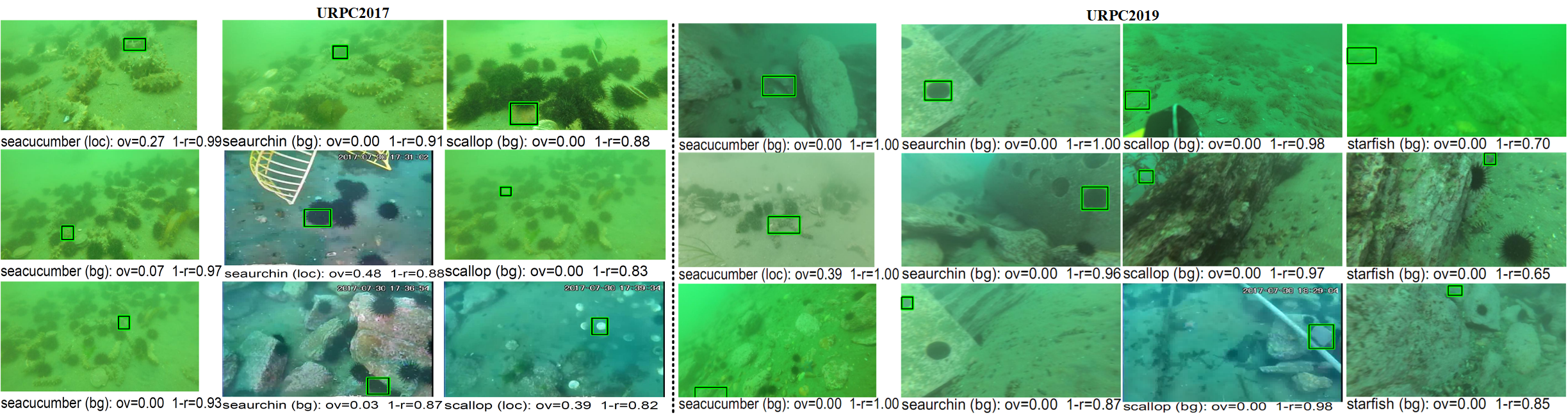}}
\caption{Examples of top false positives of SWIPENET without CMA: We show the top three false positives (FPs) for all categories on URPC2017 and URPC2019. The text indicates the type of error ("loc"=localization; "bg"=confusion with backgrounds), the amount of overlap ("ov") with a true object, and the fraction of correct examples that are ranked lower than the given false positive ("1-r", for 1-recall). Localization errors are due to insufficient overlaps (less than 0.5).}
\label{fig:tpfp}
\end{figure*}
\begin{figure*}[h]
\centerline{\includegraphics[width=18cm, height=6cm]{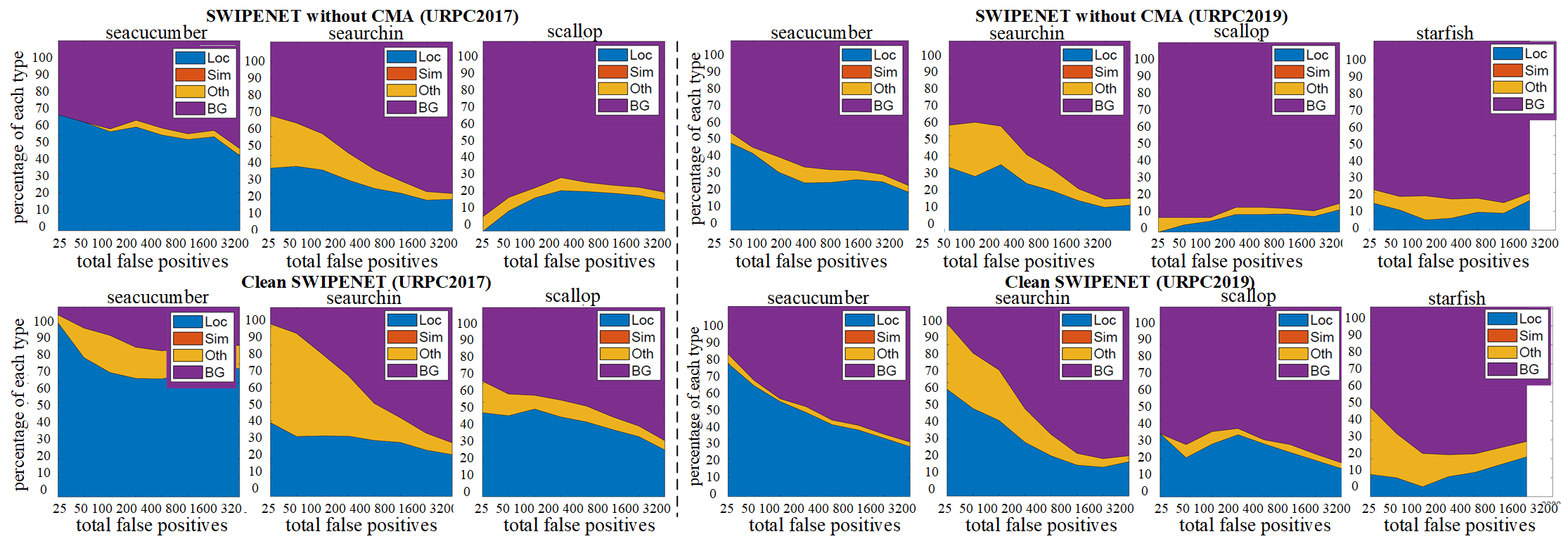}}
\caption{The distribution of top-ranked false positive types of the 1st detector in NECMA (top) and the 'clean' SWIPENET (bottom) for each category on URPC2017 and URPC2019. The false positive types include localisation error (Loc), confusion with similar categories (Sim), with others (Oth), or with background (BG).}
\label{fig:errors}
\end{figure*}
Fig.~\ref{fig:tpfp} presents the top three false positives for SWIPENET-noCMA (i.e., the SWIPENET trained without CMA) on URPC2017 and URPC2019, and Fig.~\ref{fig:errors} shows the distribution of top-ranked false positive types of SWIPENET-noCMA on URPC2017 and URPC2019. We can see that SWIPENET-noCMA makes more background errors than localisation errors. It frequently detects the backgrounds as the objects due to the influence of the noisy data.

\clearpage
\begin{figure}[htbp]
\centerline{\includegraphics[width=14cm, height=6cm]{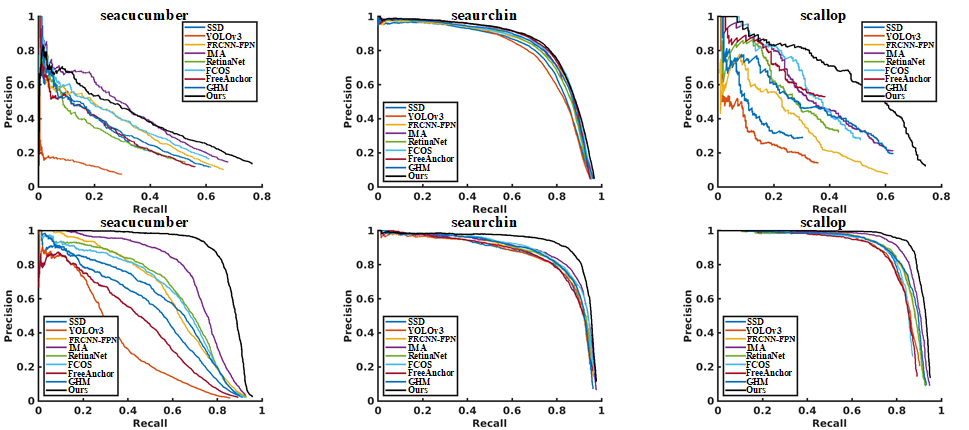}}
\caption{Precision/Recall curves of different detection methods on URPC2017 (top row) and ChinaMM (bottom row).}
\label{fig:roc17}
\end{figure}
Fig. 5 presents Precision/Recall curves of different detection methods on URPC2017 and ChinaMM, where we observe SWIPENET-CMA (black curve) achieves the best performance across all the object categories.

\begin{figure*}[htbp]
\centerline{\includegraphics[width=18cm, height=6cm]{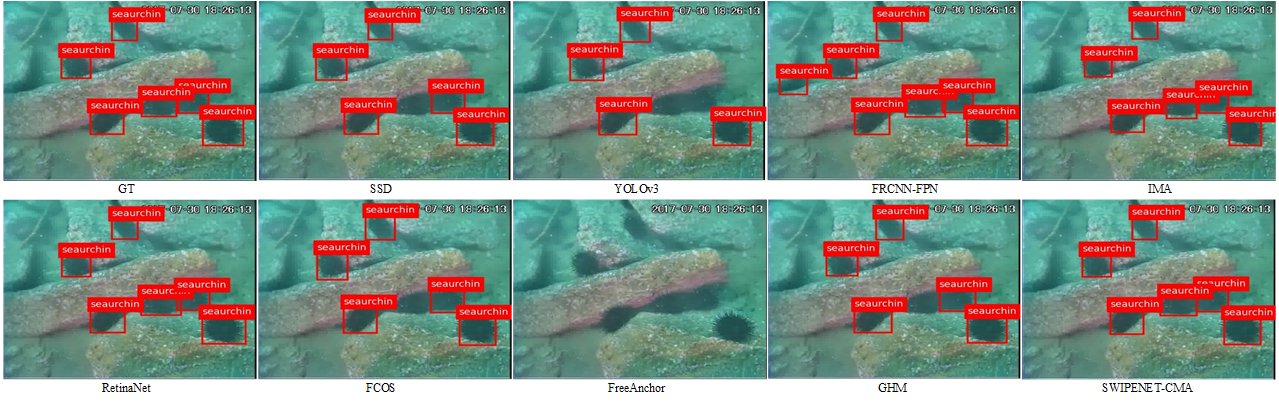}}
\caption{Visualization of object detection results of different detection frameworks on URPC2017. GT denotes the image with ground truth annotations, and the top left black box shows a starfish that looks extremely similar to the background.}
\label{fig:visualisationURPC2017}
\end{figure*}

Figs. 6 and 7 present visualization of object detection results of different detection frameworks on URPC2017 and URPC2019, most of the detection frameworks cannot detect all the objects, some of them detected the backgrounds as the objects. Among them, SWIPENET-CMA performs best.

\clearpage
\begin{figure*}[htbp]
\centerline{\includegraphics[width=18cm, height=6cm]{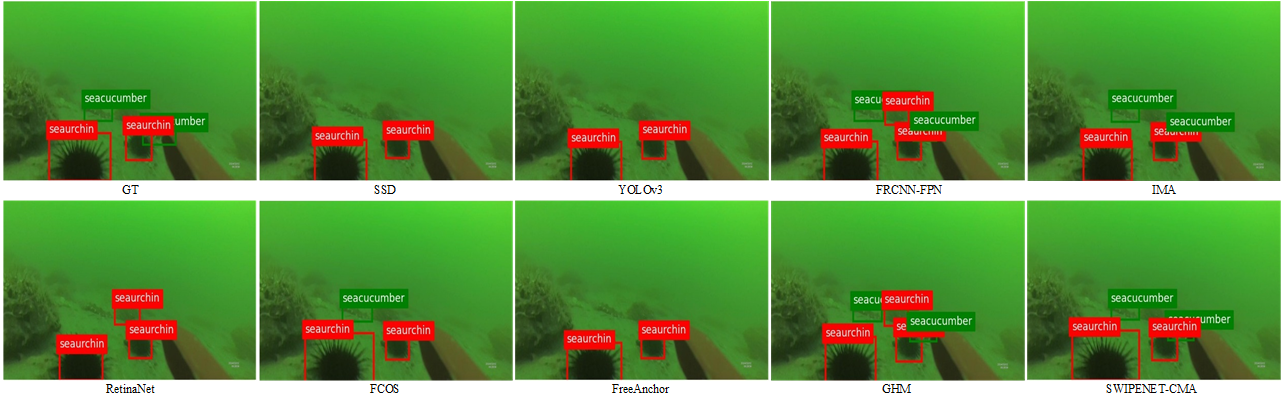}}
\caption{Visualization of object detection results of different detection frameworks on UPRC2019.}
\label{fig:visualisationURPC2019}
\end{figure*}

\subsection{The official leaderboard of the URPC competition}
\textcolor{black}{The leaderboard is published to all the teams who have attended the URPC competion. Table II shows the official leaderboard of the URPC2017 competition, which is an anonymous leaderboard with mean Average Precision (mAP). URPC2018 and URPC2019 competition haven't published the testing sets and the official leaderboards.}
\begin{table}[htbp]\footnotesize
\begin{center}
\caption{The official leaderboard of URPC2017 competition.}
\begin{tabular}{lccccccccccc}
\hline
Method & 1 & 2 & 3 & 4 & 5 & 6 & 7 & 8 & 9\\
mAP(\%) & \textbf{45.1} & 35.7 & 33.4 & 32.0 & 30.2 & 29.6 & 28.8 & 28.4 & 26.6  \\
\hline
\end{tabular}
\end{center}
\label{tab:leadb}
\end{table}

\end{document}